%% file: main-icml.tex
\documentclass{article}

\usepackage{microtype}
\usepackage{graphicx}
\usepackage{subfigure}
\usepackage{booktabs} %
\usepackage{hyperref}
\usepackage{url}
\usepackage{makecell}
\usepackage{colortbl}
\usepackage[svgnames]{xcolor}
\usepackage{graphicx}
\usepackage{multirow}
\usepackage{comment}
\usepackage[most]{tcolorbox}
\usepackage{paralist}
\usepackage{enumitem}

\usepackage[accepted]{icml2025}

\usepackage{amsmath}
\usepackage{amssymb}
\usepackage{mathtools}
\usepackage{amsthm}

\usepackage[capitalize,noabbrev]{cleveref}
\crefname{section}{Sec.}{Secs.}
\crefname{algorithm}{Alg.}{Algs.}
\crefname{appendix}{App.}{Apps.}
\crefname{table}{Table}{Tables}
\usepackage{palette}

\definecolor{outrageousorange}{rgb}{1.0, 0.43, 0.29}
\hypersetup{
colorlinks=true,linkcolor=Orchid,citecolor=outrageousorange
}

\newtcolorbox[
auto counter,
crefname={Takeaway}{Takeaways}]%
{takeawaybox}[2][]{colframe=petroil2,
colback=lacamlilac!04!white,
colbacktitle=petroil2,
fonttitle=\bfseries,
arc=0pt,outer arc=0pt,
enhanced,
attach boxed title to top left={yshift=-1pt},
boxed title style={arc=0pt,outer arc=0pt},
#1}
\theoremstyle{plain}

\theoremstyle{definition}

\theoremstyle{remark}

\newcommand{\varTarget}{T}
\newcommand{\varQuantified}{V}
\newcommand{\anchorA}{a}
\newcommand{\relR}{r}

\usepackage{xspace}
\newcommand*{\eg}{e.g.\@\xspace}
\newcommand*{\ie}{i.e.\@\xspace}

\newcommand*{\fbnew}{{\footnotesize$\mathsf{FB15k237}$}\xspace}
\newcommand*{\fb}{{\footnotesize$\mathsf{FB15k}$}\xspace}
\newcommand*{\nell}{{\footnotesize$\mathsf{NELL995}$}\xspace}

\newcommand*{\fbnewH}{{\footnotesize$\mathsf{FB15k237}$+$\mathsf{H}$}\xspace}
\newcommand*{\nellH}{{\footnotesize$\mathsf{NELL995}$+$\mathsf{H}$}\xspace}

\newcommand*{\icews}{{\footnotesize$\mathsf{ICEWS18}$}\xspace}

\newcommand*{\icewsH}{{\footnotesize$\mathsf{ICEWS18}$+$\mathsf{H}$}\xspace}

\newcommand*{\icewsCQ}{{\footnotesize$\mathsf{ICEWS18}$+$\mathsf{CQ}$}\xspace}

\newcommand{\GraphTrain}{G_{\operatorname{train}}}
\newcommand{\GraphTest}{G_{\operatorname{test}}}

\icmltitlerunning{Is Complex Query Answering Really Complex?}

\begin{document}
\twocolumn[
\icmltitle{Is Complex Query Answering Really Complex?\textsuperscript{*}}
\icmlsetsymbol{equal_super}{\includegraphics[width=.015\textwidth]{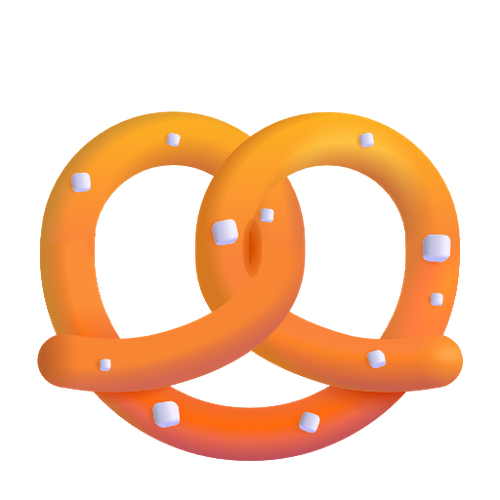}}
\begin{icmlauthorlist}
\icmlauthor{Cosimo Gregucci}{stutt}
\icmlauthor{Bo Xiong}{sta}
\icmlauthor{Daniel Hernández}{stutt}
\icmlauthor{Lorenzo Loconte}{edi}\\
\icmlauthor{Pasquale Minervini}{edi,mini,equal_super}
\icmlauthor{Steffen Staab}{stutt,south,equal_super}
\icmlauthor{Antonio Vergari}{edi,equal_super}
\end{icmlauthorlist}

\icmlaffiliation{stutt}{Institute for Artificial Intelligence, University of Stuttgart, Germany}
\icmlaffiliation{edi}{School of Informatics
University of Edinburgh
Edinburgh, UK }
\icmlaffiliation{south}{University of Southampton, UK}
\icmlaffiliation{sta}{Stanford University}
\icmlaffiliation{mini}{Miniml.AI}

\icmlcorrespondingauthor{Cosimo Gregucci}{cosimo.gregucci@ki.uni-stuttgart.de}

\icmlkeywords{complex query answering, knowledge graphs, multi-hop reasoning, neuro-symbolic}

\vskip 0.3in
]

\printAffiliationsAndNotice{\icmlSharedSupervision. \textsuperscript{*}This version differs from the ICML 2025 proceedings version: it fixes typos and aligns all query types including negation to the same filter of trivial answers, which changes some of their results (see \cref{app:easy-answers}). All claims of the paper remain unchanged.}

\begin{abstract}

Complex query answering (CQA) on knowledge graphs (KGs) is gaining momentum as a challenging reasoning task.
In this paper, we show that the current benchmarks for CQA might not be as \textit{complex} as we think, as the way they are built distorts our perception of progress in this field.
For example, we find that in these benchmarks, most queries (up to 98\% for some query types) can be reduced to simpler problems, \eg link prediction, where only one link needs to be predicted.%
The performance of state-of-the-art CQA models decreases significantly when such models are evaluated on queries that cannot be reduced to easier types.
Thus, we propose a set of more challenging benchmarks composed of queries that \textit{require} models to reason over multiple hops and better reflect the construction of real-world KGs.
In a systematic empirical investigation, the new benchmarks show that current methods leave much to be desired from current CQA methods.
\end{abstract}
\section{Introduction}

\input{./introduction}

\section{What is the real ``Hardness'' of CQA in incomplete KGs?}
\label{sec:hardness}
\input{hardness}

\section{How Many ``Easy'' Queries in Current CQA Benchmarks?}\label{sec:inv_exst_benchs}
\input{inv_exst_benchs}

\section{What do SoTA Models Perceive as Hard?}\label{sec:performance}
\input{performance}

\section{New Benchmarks for Undistorted CQA Performance}\label{sec:new-benchmarks}
\input{new-benchmarks}

\section{Conclusion}

In this paper, we revisit CQA on KGs and reveal that the ``good'' performance of SoTA approaches predominantly comes from answers that can be reduced to easier types (\cref{tab:baselines_old_bench}), the vast majority of which boiling down to single link prediction (\cref{tab:old-query-answers-type-stats}).
We perform such an analysis by dissecting the arbitrary assumptions that were used to create these datasets and by highlighting how current neural and hybrid solvers can exploit (different) forms of triple memorization to make complex queries much easier.
We confirm this by reporting their performance in a stratified analysis and by proposing our hybrid solver, CQD-Hybrid, which, while being a simple extension of an old method like CQD, can be very competitive against other SoTA models.
This reveals that \fbnew and \nell are not suitable to precisely assess the capability of CQA methods to answer complex queries{ and that most of their performance of CQA on \fbnew and \nell can be explained by their memorization ability.}

We then created a set of new benchmarks \fbnewH,\ \nellH and \icewsH that comprise the same amount of full-inference and partial-inference QA pairs. { The MRR performance of existing CQA greatly drops on our new benchmarks.}
We consider {our new benchmarks} to be a stepping stone towards benchmarking truly complex reasoning with ML models and advise all future works on CQA to use them. 
However, both the old and our new benchmarks only consider queries with bindings of single target variables.
While this reflects an important class of queries on real-world KGs, many real-world queries require bindings to multiple target variables (\ie answer tuples).
We plan to extend our current study of the ``real hardness'' of CQA benchmarks to other popular settings in neural query answering, such as inductive scenarios \citep{galkin2022inductive,arun2025semmasemanticawareknowledge} where some entities and/or relations are unseen during the test stage. { Moreover, our work could also be extended to queries requiring bindings to multiple target variables \citep{yin2023text}, and to queries having a DAG structure \citep{he2025dage}}.

\section*{Impact Statement}
This work analyses current CQA benchmarks for knowledge graphs and shows that the way benchmarks are created inflates perceived progress in the field.
By demonstrating that most benchmark queries can be reduced to simpler problems and developing new, more balanced benchmarks, this work helps establish more accurate measures of model performance in CQA tasks.
The findings have important implications for how the research community evaluates and develops CQA systems, potentially leading to more robust and capable models to better handle truly complex queries.
\section*{Author Contributions}
AV, CG, and LL conceived the initial idea for which existing benchmarks were not hard enough for complex query answering. 
AV, CG, DH, and LL had the intuition that a simple hybrid solver could obtain SoTA results on the old benchmarks. 
BX, and CG found the issue with union queries. CG developed the necessary code, ran all the experiments, and plotted the figures, with the following exceptions: BX ran all the experiments involving ConE, LL contributed to the generation of the new benchmarks, specifically for the 4p and 4i queries, PM wrote the script for generating the KG splits of ICEWS18, AV drew Fig.1 and 2. 
BX, and CG wrote the initial draft of the paper, while AV shaped the story line of the paper and helped with the writing.
All authors critically revised the paper.
AV, PM, and SS supervised all the phases of the project and gave feedback. 

\section*{Acknowledgements}
AV was supported by the ``UNREAL: Unified Reasoning Layer for Trustworthy ML'' project (EP/Y023838/1) selected by the ERC and funded by UKRI EPSRC.
PM was partially funded by ELIAI (The Edinburgh Laboratory for Integrated Artificial Intelligence), EPSRC (grant no.\ EP/W002876/1), an industry grant from Cisco, and a donation from Accenture LLP. 
CG was funded by the CHIPS Joint Undertaking (JU) under grant agreement No. 101140087 (SMARTY), and by German Federal Ministry of Education and Research (BMBF) under the sub-project with the funding number 16MEE0444.
CG acknowledge compute time on HoreKa HPC (NHR@KIT), funded by the BMBF and Baden-Württemberg’s MWK through the NHR program, with additional support from the DFG. 
BX, DH were supported by the 
Deutsche Forschungsgemeinschaft (DFG, German Research Foundation) – SFB 1574 – 471687386.
\bibliography{refs}
\bibliographystyle{icml2025}

\newpage

\input{appendix}

\end{document}

%% file: introduction.tex
\begin{figure*}[t]
\begin{center}
    \includegraphics[width=.77\textwidth,page=1,trim=6 0 6 0,clip]{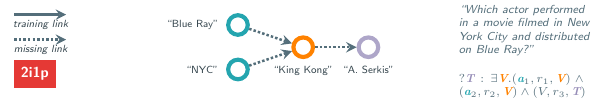}\\[-4pt]
    \par\noindent\rule{\textwidth}{0.05pt}
\end{center}
    \includegraphics[width=.32\textwidth,page=2]{figures/queries.pdf}
    \hfill
    \includegraphics[width=.32\textwidth,page=3]{figures/queries.pdf}
    \hfill
    \includegraphics[width=.32\textwidth,page=4]{figures/queries.pdf}
    \caption{
    \textbf{Query answers are not all equally hard when some links can be found in the training data} as shown for the  2i1p query \ref{query2i1pexample} and fragments of the KG \fbnew, where $r_1=\mathsf{distributedVia}^{-1}$, $r_2=\mathsf{locatedIn}^{-1}$ and $r_3=\mathsf{performedIn}^{-1}$. 
    Its different answers can be obtained by traversing the training graph (continuous line) and predicting the missing links (dotted lines).  
    (Top) Example answer that requires all links to be predicted.
    (Bottom) Example answers that require only a subset of the links to be predicted and that, therefore, can be reduced to the simpler types \ref{1p}, \ref{2p}, and \ref{2i} (see \cref{sec:background} and \cref{fig:query_types}).}
    \label{fig:2i1p-query-reduction}
\end{figure*}

A crucial challenge in AI and ML is learning to perform \textit{complex reasoning}, \ie, solving tasks that involve several intermediate steps and sub-goals to be completed.
Complex query answering~\citep[CQA;][]{DBLP:conf/nips/HamiltonBZJL18, DBLP:conf/nips/ZhangWCJW21,DBLP:conf/iclr/ArakelyanDMC21,DBLP:conf/icml/Zhu0Z022} emerged as a way to measure complex reasoning over external knowledge bases, encoded as knowledge graphs~\citep[KGs;][]{hogan2021knowledge}.
For instance, to answer the query:
\begin{equation}
\tag{$q_1$}
\begin{aligned}
    &\text{``\textit{Which actor performed in a movie filmed in}} \\
    &\text{\textit{New York City and distributed on Blue Ray?}''}
\end{aligned}
\label{query2i1pexample}
\end{equation}
\noindent over a KG such as FreeBase~\citep{DBLP:conf/sigmod/BollackerEPST08}, one 
would need to \textit{first} intersect the set of movies found on  \textit{Blue Ray} and the ones shot in \textit{New York City}, and \textit{then} link these intermediate candidate answers to another entity, i.e., an \textit{actor} participating in it.
However, if some links are missing in the KG, not all relevant actor entities can be immediately reachable.
To deal with the unavoidable incompleteness of real-world KGs, ML methods were developed to solve CQA in the presence of missing links.
The current state-of-the-art (SoTA) for CQA are \textit{neural query answering} models
 \citep{DBLP:conf/iclr/ArakelyanDMC21, DBLP:conf/icml/Zhu0Z022, DBLP:conf/nips/ArakelyanMDCA23,ren2023neural, DBLP:journals/corr/abs-2404-07198}, mapping queries and KGs (\ie, entities and relation names) into a unified latent space. 
Performance measured~on de-facto-standard benchmarks such as \fbnew~\citep{DBLP:conf/acl-cvsc/ToutanovaC15} and \nell~\citep{DBLP:conf/emnlp/XiongHW17} suggests impressive progress achieved in recent years on CQA on queries having different structures, and hence posing apparently different levels of difficulty to be answered.

The difficulty of a benchmark relates to the size and structural complexity\footnote{Not to be confused with the computational complexity of query answering \citep{DBLP:journals/jacm/DalviS12}.
} of its queries, and several query ``types'' have been proposed \citep{DBLP:conf/iclr/RenHL20}, each involving a different combination of logical operators---conjunctions, disjunctions, and requiring to traverse a number of missing links that generally increases with the number of logical conditions imposed (\cref{fig:2i1p-query-reduction,fig:query_types}).
For instance, the query \ref{query2i1pexample} is an example of a ``\textit{2i1p}'' query type, since it comprises an intersection of two entity sets \textit{(2i)} followed by a path\footnote{Also known as \textit{projection} in related works \citep{ren2023neural}.} of length one \textit{(1p}).
In this paper, we argue that \textbf{\textit{the perception of progress on CQA benchmarks has been distorted by implicit assumptions in these benchmarks}}.
We start by noting how both in \fbnew and \nell, the vast majority of queries of a complex type simplify to one simpler type, thanks to the fact that to answer them, one can leverage links already appearing in the training data.
\cref{fig:2i1p-query-reduction} shows an example of how answers to the query \ref{query2i1pexample} on \fbnew -- which should require predicting \textit{three} missing links -- require the same effort associated with simpler types involving fewer links.
In fact, in \fbnew, the answer ``K. Dunst'' can be retrieved by predicting just one missing link when leveraging the training links.
In this sense, a 2i1p query reduces to a ``1p'' query, \ie, the much simpler task of link prediction (see \cref{sec:background}).
Similarly, the answers ``J. Bryant'' and ``K. Chandler'' can be retrieved by predicting two links instead of three. 
Only the answer ``A. Serkis'' requires predicting three missing links, and thus 
follows the intuitive ``hardness'' associated to the type 2i1p, i.e., it is the hardest when no information is available. %
As we note here, there is a \textit{spectrum of hardness}, predicting ``K. Dunst'' will be easier than answering ``K.Chandler'', which in turn is easier than saying ``A. Serkis''.
However, current CQA benchmarks greatly overrepresent the easy query answers over the others -- the ``K. Dunst''-like answers are the vast majority -- and this inflates performance as SoTA models are known to be prone to memorize training links.
In this paper, we analyze this issue and propose a new set of benchmarks that 
cover the full spectrum of hardness in CQA.

\textbf{Contributions.}
After revisiting the CQA task (\cref{sec:background}) and highlighting how query types can simplify at test time in the presence of training links (\cref{sec:hardness}),
(\textbf{C1}) we show that up to 99.9\% of the test queries can be reduced to simpler queries, with the majority of them (up to 98\%) reduced to ``one-step'' link prediction problems (\cref{sec:inv_exst_benchs}).
This clearly inflates the reported performance.
(\textbf{C2}) We re-evaluate previous SoTA approaches (\cref{sec:performance}), revealing that neural link predictors rely on memorized information from the training set.
Furthermore, we show that the reported hardness of queries involving unions is only apparent.
To better understand why performances are inflated, we introduce CQD-Hybrid, a hybrid CQA solver that combines classical graph matching with neural link predictors.
(\textbf{C3}) We create new benchmarks for CQA (\cref{sec:new-benchmarks}), \fbnewH and \nellH from \fbnew and \nell, respectively, where we equally sample all intermediate query types one can consider by accounting for known links in the training data.
To these, we build \icewsH from the temporal KG \icews \citep{DVN/28075_2015}, where validation and test links are links that have been added to the KG \textit{after} those in the training, to investigate more realistic sampling schemes to generate the train/validation/test splits.
Lastly, we add the query types ``4p'' (a four-length path) and ``4i'' (a conjunction of four patterns), making them more challenging.

\section{KGs and Complex Query Answering}
\label{sec:background}

\textbf{Knowledge graphs.}
A KG is a graph-structured knowledge base where information is encoded as relationships between entities.
More formally, a KG is a multi-relational graph $G=(\mathcal{E}, \mathcal{R}, \mathcal{T})$, where $\mathcal{E}$ is a set of entities, $\mathcal{R}$ is a set of relation names, and $\mathcal{T} \subseteq \mathcal{E} \times \mathcal{R} \times \mathcal{E}$ is a set of links or \emph{triples}, where each triple $(s, p, o) \in \mathcal{T}$ represents a relationship of type $p$ between a subject $s$ and an object $o$. 
For instance, in a KG such as FreeBase, the fact that the movie ``Spiderman 2'' is distributed in Blue Ray is stored in the triple $(\mathsf{Spiderman 2}, \mathsf{distributedVia}, \mathsf{Blue Ray})$ or equivalently $(\mathsf{Blue Ray}, \mathsf{distributedVia}^{-1}, \mathsf{Spiderman 2})$ where $\mathsf{distributedVia}^{-1}$ is the inverse relation of $\mathsf{distributedVia}$.
\cref{fig:2i1p-query-reduction} shows examples of fragments of the KG \fbnew.

\textbf{The aim of CQA} is to retrieve a set of answers to a logical query $q$ that poses conditions over entities and relation types in a KG.
Following~\citet{DBLP:conf/iclr/ArakelyanDMC21}, we consider the problem of answering logical queries with a single \textit{target} variable ($T$), a set of constants including \textit{anchor} entities ($a_1,\ldots,a_k\in\mathcal{E}$), given relation names ($r_1,\ldots,r_n\in\mathcal{R}$), and first-order logical operations that include conjunction $\wedge$, disjunction $\vee$, negation $\neg$ and existential quantification $\exists$. 
\begin{figure*}[!t]
    \centering
    \begin{tabular}{cccccc}
         {\includegraphics[height=.06\textwidth,page=5]{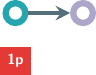}}
         & 
         {\includegraphics[height=.06\textwidth,page=6]{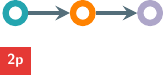}}
         & 
         \includegraphics[height=.06\textwidth,page=7]{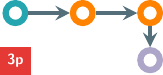}
         & 
         \includegraphics[height=.06\textwidth,page=8]{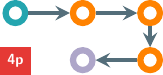}
         & 
          \includegraphics[height=.06\textwidth,page=9]{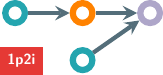}
         &
         \includegraphics[height=.06\textwidth,page=10]{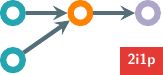}
         \\[5pt] 
         \includegraphics[height=.06\textwidth,page=11]{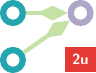}
         & 
         \includegraphics[height=.06\textwidth,page=12]{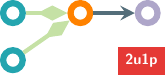}
         & 
         \includegraphics[height=.06\textwidth,page=13]{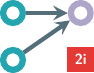}
         & 
          \includegraphics[height=.06\textwidth,page=14]{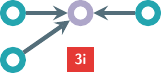}
          & 
          \includegraphics[height=.06\textwidth,page=15]{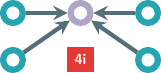}
          & 
          \raisebox{-4pt}{\includegraphics[height=.07\textwidth,page=16]{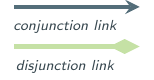}}
         \\
    \end{tabular}
    \caption{\textbf{Query structures we consider}, adapted from \citet{DBLP:conf/nips/RenL20} and including \textit{path} ({\bf p}), \textit{intersection} ({\bf i}), \textit{union} ({\bf u}) structures. \cref{sec:background} for their logical formulation.
    See \cref{app:negation_queries} for additional query structures including \textbf{negation}.}
    \label{fig:query_types}
\end{figure*}
Different queries are categorized into different \textit{types} based on the structure of their corresponding logical sentence \citep{DBLP:conf/emnlp/XiongHW17}.
This taxonomy should imply that queries of the same type share the same ``hardness'', \ie, the level of difficulty to be answered,
and different types correspond to tasks that map to more or less complex reasoning tasks.
As we will show in the next sections, this is not the case, and to fully grasp the actual hardness to answer a query one needs to account for the ``leakage'' of intermediate training links.
We now review the most common query types used in CQA.
The simplest CQA task is \textit{link prediction}, \ie, answering a query of the form:
\begin{equation}  \label{1p}
    \tag{1p} ?\varTarget:(\anchorA_1, \relR_1, \varTarget),
\end{equation}
\noindent that is, given an entity $a_1$ (\eg, $\mathsf{NYC}$) and a relation name $r_1$ (e.g., $\mathsf{locatedIn}^{-1}$), find the entity that when substituted to $T$ correctly matches the link in the KG (e.g., $\mathsf{Spiderman 2}$, see \cref{fig:2i1p-query-reduction}). 
Instead of matching a single link, more complex queries involve matching \textit{sub-graphs} in a KG (see \cref{fig:query_types}).
\citet{DBLP:conf/emnlp/XiongHW17} extend \ref{1p} queries,  and ask questions involving sequential \textit{paths} made of two or three links, i.e., 
\begin{align} \label{2p}
    \tag{2p} &\exists \varQuantified_1 . (\anchorA_1, \relR_1, \varQuantified_1) \land (\varQuantified_1, \relR_2, \varTarget),\\
    \tag{3p} &\exists \varQuantified_1, \varQuantified_2 . (\anchorA_1, \relR_1, \varQuantified_1) \land (\varQuantified_1, \relR_2, \varQuantified_2) \land (\varQuantified_2, \relR_3, \varTarget),\label{3p}
\end{align}
\noindent where $\varQuantified_1,\varQuantified_2$ denote variables to be grounded into entities appearing in the path.
Moreover, multiple ground entities can participate in a conjunction, e.g., in queries such as:
\begin{align}
\label{2i}
\tag{2i} &?T:(\anchorA_1, \relR_1, \varTarget) \land (\anchorA_2, \relR_2, \varTarget),\\
\tag{3i} &?T:(\anchorA_1, \relR_1, \varTarget) \land
    (\anchorA_2, \relR_2, \varTarget) \land
    (\anchorA_3, \relR_3, \varTarget),\label{3i}
\end{align} 
which represent the \textit{intersection} of the target entity sets defined over two (\ref{2i}) or three (\ref{3i}) links. Path and intersection structures can be combined into more complex queries: for example, the natural language expression for query \ref{query2i1pexample} can be formalized as the formula
\begin{equation} \label{2i1p}
\tag{2i1p} ?T:\exists \varQuantified_1 .
    (\anchorA_1, \relR_1, \varQuantified_1) \land
    (\anchorA_2, \relR_2, \varQuantified_1) \land
    (\varQuantified_1, \relR_3, \varTarget),
\end{equation}
\noindent involving one intersection followed by a length-one path.\footnote{In \citet{DBLP:conf/iclr/RenHL20} this type was referred to as ``ip''.
We explicitly mention the number of steps involved in a path or conjunction, as this is a factor of complexity. Analogously, ``pi'', and ``up'' queries from \citet{DBLP:conf/iclr/RenHL20} are now 1p2i, and 2u1p.}
See also the example in \cref{fig:2i1p-query-reduction}. 
By inverting the order of operations, we obtain the query type ``1p2i'':
\begin{equation} \label{1p2i}
\tag{1p2i} ?T:\exists \varQuantified_1 .
    (\anchorA_1, \relR_1, \varQuantified_1) \land
    (\varQuantified_1, \relR_2, \varTarget) \land
    (\anchorA_2, \relR_3, \varTarget).
\end{equation}
Similarly to introducing conjunctions, we can consider disjunctions in queries, realizing the \textit{union} query types which can be answered by matching one link or the other, such as 
\begin{equation} \label{2u}
\tag{2u} ?T:(\anchorA_1, \relR_1, \varTarget) \lor (\anchorA_2, \relR_2, \varTarget),
\end{equation}
or by combining the union with the previous query types, e.g., obtained by combining \ref{2u} and \ref{1p}:
\begin{equation} \label{2u1p}
\tag{2u1p} ?T:\exists \varQuantified_1 . ((\anchorA_1, \relR_1, \varQuantified_1) \lor (\anchorA_2, \relR_2, \varQuantified_1)) \land (\varQuantified_1, \relR_2, \varTarget).
\end{equation}
Note that despite their dissimilar syntaxes and the associated sub-graphs (\cref{fig:query_types}), \textit{the query type \ref{2u} should be as difficult as \ref{1p}}, as to answer the first it suffices to match a single link correctly.
Similarly, \ref{2u1p} should be as complex as \ref{2p}. The fact that \ref{2u} and \ref{2u1p} are reported to be harder to solve in practice than \ref{1p} and \ref{2p} \citep{DBLP:conf/iclr/RenHL20} is due to the way standard benchmarks are built, which we discuss in \cref{sec:inv_exst_benchs}, and the way in which CQA is evaluated, discussed next.
\cref{app:negation_queries} details further query types, including negation.

\textbf{Standard evaluation.}
Given a KG $G$ and a query $q$, answering $q$ boils down to a graph matching problem \citep{hogan2020sparql} if we assume that all the meaningful links are already in $G$.
Instead, if $G$ is incomplete, we will need to predict missing links while answering $q$.
Many ML approaches to CQA, reviewed in \cref{sec:performance}, therefore assume a distribution over \textit{possible} links \citep{NEURIPS2023_f4b76818}, thus requiring some form of \textit{probabilistic reasoning}.
To evaluate them, standard benchmarks such as \fbnew and \nell artificially divide $G$ into $\GraphTrain$ and $\GraphTest$, treating the triples in the latter as missing links.
This splitting process is done uniformly at random, a procedural choice that is inherited from link prediction benchmarks \citep{DBLP:conf/iclr/RuffinelliBG20} and that can alter the measured performance, as we discuss next.
CQA is generally treated as a ranking task, counting how many true candidate answers to a query are ranked higher than non-answer ones.
Let the rank of a \textbf{\textit{query answer (QA) pair}} $(q,t)$ be $\operatorname{rank}(q,t)$,  the performance for each query type is calculated as the mean reciprocal rank (MRR), \ie:
\begin{equation} \label{eq_metrics}
     |\mathcal{Q}|^{-1} \sum\nolimits_{q \in \mathcal{Q}, t \in \mathcal{E}_q } 
     \left| \mathcal{E}_q \right|^{-1}
     \operatorname{rank}(q,t)^{-1},
\end{equation}
where $\mathcal{Q}$ denotes the set of test queries of the considered type, and $\mathcal{E}_q$ is the set of candidate answer entities for each query $q\in\mathcal{Q}$.
This average, across queries and answers, assumes that every QA pair having the same query type is equivalently hard, which is not the case.
In fact, we show that certain QA pairs can be easier to predict if some links were already seen by the model during training (\cref{fig:2i1p-query-reduction}) and that the distribution of the query answer pairs in the existing benchmarks is very skewed towards those involving a single missing link (\cref{tab:old-query-answers-type-stats}).
Thus, computing \cref{eq_metrics} without understanding what the benchmark distribution is distorts the perception of performance gains.

%% file: hardness.tex
As discussed in the previous section, the perceived complexity of a query is related to the graph structure associated to its query type (\cref{fig:query_types}): queries containing more hops/existentially-quantified variables are more challenging, e.g., a \ref{3p} query is harder than a \ref{2p} query.
In this section, we give an alternative perspective on the difficulty of answering queries that take into account the information coming from the training data. 
We argue that predicting links that are truly missing, i.e., not accessible to a learned model, is actually what makes a query ``hard''.
To do so, we formalize the notion of a reasoning tree for a QA pair, and then define how we determine the practical hardness of a QA pair.

\begin{table*}[!th]
\centering
\small
\setlength{\tabcolsep}{5pt} %
\caption{
\textbf{The great majority of QA pairs are of the much easier partial-inference type (non-diagonal), rather than of the harder full-inference type (diagonal).}
For each query type (as rows)
we show
the percentage of the QA pairs that can be reduced to an easier query type (as columns) for datasets \fbnew and \nell.
Most of the complex queries can be reduced to simple link prediction queries (i.e., \ref{1p}).
We denote as `-' those reductions that are not possible
given the sub-graph structure induced by the query type.
}
\vspace{4pt}
\setlength{\tabcolsep}{4pt}
\scalebox{0.95}{
\begin{tabular}{rrrrrrrrrrrrrrrrrrr}
\toprule
& \multicolumn{9}{c}{\fbnew} & \multicolumn{9}{c}{\nell} \\
\cmidrule(lr){2-10} \cmidrule(lr){11-19}
& 1p & 2p & 3p & 2i & 3i & 1p2i & 2i1p & 2u & 2u1p & 1p & 2p & 3p & 2i & 3i & 1p2i & 2i1p & 2u & 2u1p \\
\midrule
1p  & \cellcolor{orange!50} 100 & - & - & - & - & - & - & - & - & \cellcolor{orange!50} 100 & - & - & - & - & - & - & - & - \\
2p  & 98.1 & \cellcolor{orange!50} 1.9 & - & - & - & - &  - & - & - & 97.6 & \cellcolor{orange!50} 2.4 & - & - & - & - & - & - & - \\
3p  & 97.2 & 2.7 & \cellcolor{orange!50} 0.1 & - & - & - & - & - & - & 95.6 & 4.3 & \cellcolor{orange!50} 0.1 & - & - & - & - & - & - \\
2i  & 96.0 & - & - & \cellcolor{orange!50} 4.0 & - & - & - & - & - & 94.0 & - & - & \cellcolor{orange!50} 6.0 & - & - & - & - & - \\
3i  & 91.6 & - & - & 8.2 & \cellcolor{orange!50} 0.2 & - & - & - & - & 87.4 & - & - & 12.1 & \cellcolor{orange!50} 0.5 & - & - & - & - \\
1p2i  & 86.8 & 1.0 & - & 12.0 & - & \cellcolor{orange!50} 0.2 & - & - & - & 49.5 & 0.6 & - & 49.0 & - & \cellcolor{orange!50} 0.9 & - & - & - \\
2i1p  & 96.7 & 1.8 & - & 1.4 & - & - & \cellcolor{orange!50} 0.1 & - & - & 96.2 & 2.4 & - & 1.2 & - & - & \cellcolor{orange!50} 0.2 & - & - \\
2u  & 0.0 & - & - & - & - & - & - & \cellcolor{orange!50} 100 &- & 0.0 &  - & - & - & - & - & - & \cellcolor{orange!50} 100 & - \\
2u1p  & 98.3 & 0.0 & - & - & - & - & - & 1.6 & \cellcolor{orange!50} 0.1 & 98.5 & 0.0 & - & - & - & - & - & 1.4 & \cellcolor{orange!50} 0.1 \\
\bottomrule
\end{tabular}
}
\label{tab:old-query-answers-type-stats}
\end{table*}

Given a query $q$ and every answer set of candidate answers $t\in\mathcal{E}_q$,
we define the
\textit{\textbf{reasoning tree}} of each QA pair $(q,t)$, as the directed acyclic graph starting from the anchor entities of $q$ to the target entity $t$, whose relational structure matches the query graph.
\cref{fig:2i1p-query-reduction} provides examples of different reasoning trees for four different answers to the same query.
There, we highlight whether a link belongs to $G_{\text{train}}$ or not, i.e., it is a missing link.
We assess the hardness of each answer $t\in\mathcal{E}_q$, by analyzing the composition of the reasoning tree required to predict $t$.
As to answering a $(q,t)$ pair, multiple reasoning trees are possible, so we select the one with the smallest number of missing links and the fewest number of hops.
A $(q,t)$ pair is \textit{\textbf{trivial}} if the answer $t$ can be entirely retrieved from $G_{\text{train}}$, \ie, there exists at least one reasoning tree where all the links in it are seen during training. 
This type of answer does not need probabilistic inference and hence is filtered out from current CQA benchmarks~\citep{DBLP:conf/nips/RenL20}, which only consider non-trivial $(q,t)$ pairs, which we call \textbf{\textit{inference-based}} pairs.
However, inference-based pairs do not need all links in their reasoning tree to be predicted as, by definition, it is sufficient that at least one link in the tree is missing.
Therefore, we define a $(q,t)$ pair to be a \textbf{\textit{partial-inference}} pair if its reasoning tree contains at least one link in $G_{\text{train}}$ and at least one link present in $G_{\text{test}}$.
Alternatively, a $(q,t)$ pair whose reasoning tree contains only links present in $G_{\text{test}}$ is called a \textbf{\textit{full-inference}} pair.
\textbf{\textit{We claim that inference-based queries encompass a spectrum of hardness, with full-inference queries being the hardest and queries that have only one truly missing link (\ref{1p}) the easiest.
However, current benchmarks flatten this complexity and highlight only performance on the easiest}}.

To predict a partial-inference pair $(q, t)$, a ML model that has \textit{\textbf{explicit}} access to $\GraphTrain$ has to solve a simpler task to answer $q$, as only a subset of the links in the reasoning tree are missing and need to be predicted.
As such, the $(q,t)$ pair can be simplified to $(q',t)$ pair, where the query $q'$ is of an easier query type than $q$. 
\cref{fig:2i1p-query-reduction} shows one example of how a QA pairs of type \ref{2i1p} from \fbnew are reduced in practice to the simpler types \ref{1p}, \ref{2i} and \ref{2p}.
This is exploited by hybrid solvers such as QTO \citep{QTO} and our CQD-Hybrid ( \cref{sec:performance}).
Note that this advantage applies also to ML models that have \textbf{\textit{implicit}} access  to $\GraphTrain$, \eg, by having memorized the triples during training, a common phenomenon for many neural link predictors~\citep{DBLP:conf/nips/NickelJT14}.
We confirm this in \cref{sec:performance}.

%% file: inv_exst_benchs.tex
In this section, we systematically analyze the practical hardness of queries from the current CQA benchmarks and answer the following research question:
\textbf{(RQ1)} \textit{What is the proportion of QA pairs that can be classified as full-inference rather than partial-inference and how easy are the latter?}
To this end, we consider the CQA benchmarks generated from \fbnew (based on Freebase)
and \nell (based on NELL systems~\citep{DBLP:conf/aaai/CarlsonBKSHM10})
as they are the most used to evaluate SoTA methods for CQA \citep{DBLP:conf/iclr/RenHL20, DBLP:conf/nips/RenL20, DBLP:conf/iclr/ArakelyanDMC21, DBLP:conf/nips/ZhangWCJW21, DBLP:conf/icml/Zhu0Z022, DBLP:conf/nips/ArakelyanMDCA23}.\footnote{We do not consider \fb \citep{DBLP:conf/nips/BordesUGWY13} as it suffers from another form of data leakage~\citep{DBLP:conf/acl-cvsc/ToutanovaC15}.}

\textbf{Complex queries can be reduced to much simpler types.}
We group the testing QA pairs into query types (\cref{sec:background}), and we further split them based on whether they can be reduced to simpler types after observing the training links in $\GraphTrain$ (\cref{sec:hardness}).
\cref{tab:old-query-answers-type-stats} shows that for both \fbnew and \nell
the vast majority of QA pairs can be reduced to simpler types.
For \fbnew, 86.8\% to 98.3\% of QA pairs can be reduced to \ref{1p} queries, while only 0.1\% to 4\% require full inference.
Similarly, for \nell, 49.5\% to 98.5\% of QA pairs map to \ref{1p} queries, and only 0.1\% to 6\% to full inference.
For instance, 96.7\% of \ref{2i1p} QA pairs in \fbnew can be reduced to \ref{1p} queries, 1.8\% to \ref{2p}, and 1.4\% to \ref{2i}.
However, only 0.1\% of these pairs cannot be reduced to any other QA pair, i.e., they require full inference in order to be predicted.
The only exceptions to this trend are QA pairs where the query has a \ref{1p} or \ref{2u} structure which, by definition, only require the prediction of a single link and therefore cannot be reduced by any other query type.
We conduct the same analysis for queries that include negation, considering only the  reasoning tree over non-negated links (see \cref{app:an_neg} for details). \cref{tab:neg-analysis} presents results analogous to those in \cref{tab:old-query-answers-type-stats}, showing that the vast majority of QA pairs including negation have training links in their reasoning tree.
For example, 95.4\% and 93.9\% of \ref{3in} QA pairs in \fbnew and \nell, respectively, have training links in their reasoning tree.

\textbf{Is this realistic?} This skewed distribution is the result of \textit{implicit assumptions} behind how these benchmarks have been created.
We argue they are arbitrary and not realistic.
First, we note that the procedure to generate queries assumes that triples are \textit{independent}, a simplifying assumption that does not hold in practice.
Second, as triples are sampled independently, the quantity of QA pairs that can be reduced to \ref{1p} depends on how many links are assumed to be missing. 
This is given by the ratio of $|\GraphTrain|$ and $|\GraphTest|$. Note that however the train/test split is arbitrary and has been inherited from the benchmarks for link prediction, where the task was rightfully to predict a single link.

\begin{figure}[!t]
\centering
\includegraphics[width=\columnwidth,page=1]{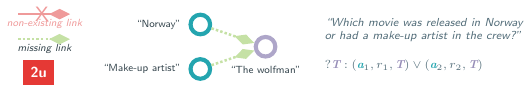}\\[-5pt]
\rule{\columnwidth}{0.05pt}
\vspace{5pt}

\begin{minipage}{0.48\columnwidth}
    \includegraphics[width=\linewidth,page=2]{figures/queries-union.pdf}
\end{minipage}
\hfill
\begin{minipage}{0.48\columnwidth}
    \includegraphics[width=\linewidth,page=3]{figures/queries-union.pdf}
\end{minipage}

\vspace{4pt}
\caption{ \small\textbf{Some answers of union queries can be reached by a single missing link, while the other link does not exist}, as shown for this 2u query and fragments of the KG \fbnew, where $r_1=\mathsf{releasedIn}$, $r_2=\mathsf{crewMemberOf}$. (Top) Example of an answer that we retain in our benchmark, being reachable by two missing links. (Bottom) Example of answers we filter out, being only reachable by one missing link.}
\label{fig:2u-query-reduction}
\end{figure}

\textbf{Non-existing links for union queries.}
Finally, we discovered that there are \textit{non-existing} links, i.e., links that are not in the original KG $G$ and hence neither in $\GraphTrain$ or $\GraphTest$, in both \fbnew and \nell  in the reasoning trees of queries involving unions, i.e., the \ref{2u} and \ref{2u1p} types.
These links violate our definitions of inference QA pairs, hence, we filter them out, and report only \textit{filtered} QA pairs in \cref{tab:old-query-answers-type-stats}. 
{
For example, for \ref{2u} QA pairs, if either one of the two links in the reasoning tree does not exist in the graph, we remove the QA pair (see \cref{fig:2u-query-reduction} bottom for an example). It follows that only QA pairs where both links in the reasoning tree exist in the test graph are retained (see \cref{fig:2u-query-reduction} (Top) for an example).
}
More crucially, these non-existing links can alter the performance of solvers for \ref{2u} and \ref{2u1p} types, as we discuss next.

\begin{takeawaybox}[label={takeaway:benchmarks}]

\textbf{Takeaway 1.}
    We discourage using current CQA benchmarks, %
    as they essentially reflect only the performance of predicting answers where only one link is truly missing (for both hybrid and neural solvers). 
\end{takeawaybox}

\begin{table*}[!t]
\centering
\setlength{\tabcolsep}{4pt}

\begin{minipage}{.67\textwidth}
    \scalebox{0.76}{
\begin{tabular}{rrrrrrrrrrrrrrrrrr}
\toprule
&\multicolumn{8}{c}{3p} && \multicolumn{8}{c}{2i1p}\\
\cmidrule(lr){2-9}\cmidrule(lr){11-18}
 method & ovr & 1p & 2p & 3p & 2i & 3i & 1p2i & 2i1p&& ovr & 1p & 2p & 3p & 2i & 3i & 1p2i & 2i1p\\
\midrule
  
  GNN-QE& 11.8 & 12.0 & 4.4 & \cellcolor{orange!50}4.8& -& -& -& -&& 
  18.9 & 19.2 & 8.2 & -& 6.2& -& -& \cellcolor{orange!50}3.4\\
  ULTRAQ& 8.9 & 9.0 & 4.6 & \cellcolor{orange!50}4.4& -& -& -& -&& 
  18.6 & 18.9 & 8.5 & -& 5.1& -& -& \cellcolor{orange!50}8.1\\
  ConE & 11.0 & 11.0 & 5.4 & \cellcolor{orange!50}2.6 & -& -& -& -& &
  14.0 & 13.8 & 9.7 & -& 12.8 & -& -& \cellcolor{orange!50}5.6\\
  CQD& 7.8 & 7.8 & 3.4 & \cellcolor{orange!50}1.8& -& -& -& -&& 
  20.7 & 21.0 & 10.5 & -& 6.7& -& -& \cellcolor{orange!50}7.6\\
  
  CQD-Hybrid& 11.0 & 11.1 & 3.6 & \cellcolor{orange!50}4.4& -& -& -& - & & 24.0 & 24.6 & 10.0 & -& 4.6& -& -& \cellcolor{orange!50}7.5\\
QTO& \textbf{15.6} & 15.8 & 4.5 & \cellcolor{orange!50}\textbf{5.0}& -& -& -& -&& \textbf{24.7} & 25.3 & 10.3 & -& 8.6& -& -& \cellcolor{orange!50}8.1\\
CLMPT& 11.3 & 11.3 & 6.1 & \cellcolor{orange!50}4.0& -& -& -& -&& 19.3 & 19.1 & 18.7 & -& 7.9& -& -& \cellcolor{orange!50}\textbf{13.2}\\

\bottomrule

\end{tabular}
}
\end{minipage}\hfill\begin{minipage}{.3\textwidth}
\vspace{-7pt}
    \caption{\textbf{SoTA performance drops significantly on full-inference QA pairs} as shown for the MRR for \ref{3p} and \ref{2i1p} queries on \fbnew. 
    Compare the overall MRR achieved on all available testing queries (column ``ovr'') with that achieved on the 1p reductions (``1p'' column, i.e., when only a single link is truly missing): they are extremely close. This is because
}
\label{tab:baselines_old_bench}
\end{minipage}
\vspace{-2pt}
\small
\flushleft
\ref{1p} queries are overrepresented in the existing benchmark (\cref{tab:old-query-answers-type-stats}).
Compare instead their values to those in the full-inference column (highlighted in orange) where all links in the reasoning tree of the QA pairs are missing: the latter are much lower.
This applies also to intermediate partial-inference QA pairs (middle columns), entailing that the skewed distribution of these datasets (\cref{tab:old-query-answers-type-stats}) greatly distorts the overall performance usually reported in previous papers.
The same pattern appears for all query types in both \fbnew and \nell.
See \cref{tab:NELL_complex query answering complex?per_answer,tab:fullcomparison?oldFB15k237} for the complete stratified analysis for all query types and models. 
    
\end{table*}

%% file: performance.tex
In this section, we re-evaluate SoTA methods for CQA and answer to the following question: \textbf{(RQ2}) \textit{How do they perform on QA pairs that are placed differently in the hardness spectrum?}
To substantiate that memorizing training links helps in answering partial-inference queries but not full-inference ones (\cref{sec:hardness}), we evaluate hybrid solvers who have access to $\GraphTrain$.

\par
\textbf{Neural CQA methods.}
Over the years, a significant number of neural models have been proposed for solving the CQA task, see \cref{app:more-related-works} for an overview.
Among purely-neural models, we consider five representative approaches that significantly differ in their methodological designs and yield SoTA results compared to other models in their class:
\begin{inparaenum}[(1)]
\item Cone Embeddings~\citep[ConE;][]{DBLP:conf/nips/ZhangWCJW21} is a geometry-based complex query answering model where logical conjunctions and disjunctions are represented as intersections and union of cones.
\item Graph Neural Network Query Executor~\citep[GNN-QE;][]{DBLP:conf/icml/Zhu0Z022} decomposes a complex first-order logic query into projections over fuzzy sets.
{
\item Conditional Logical Message Passing Transformer~\citep[CLMPT;][] {zhang2024conditional} leverages existing KG embeddings to conduct one-hop inferences on atomic formulas in a message-passing scheme as in LMPNN \citep{wang2023logical}, and extends it by using a transformer to aggregate messages incoming from neighbor nodes.
}
\item Continuous Query Decomposition~\citep[CQD;][]{DBLP:conf/iclr/ArakelyanDMC21,DBLP:conf/nips/ArakelyanMDCA23}, reduces the CQA task to the problem of finding the most likely variable assignment, where the likelihood of each link (\ref{1p} query) is assessed by a neural link predictor, and logical connectives are relaxed via fuzzy logic operators.
\item UltraQuery~\citep[ULTRAQ;][]{DBLP:journals/corr/abs-2404-07198} is a foundation model for CQA inspired by~\citet{DBLP:conf/iclr/0001YM0Z24} where links and logical operations are represented by vocabulary-independent functions which can generalize to new entities and relation types in any KG.
\end{inparaenum}

\textbf{Hybrid solvers.} So far, we have assessed that current benchmarks are skewed towards easier query types (\cref{tab:old-query-answers-type-stats}) and thus the perceived progress of current SoTA CQA methods boils down to their performance on \ref{1p} queries (\cref{tab:baselines_old_bench}).
To have an undistorted view of this progress, in the next section, we will devise a benchmark that rebalances all partial-inference as well as full-inference queries types.
That is, we will provide an equal number of QA pairs for each sub-type a query can be reduced to.
Now, we argue that in a real-world scenario one has to perform reasoning over \textit{both} existing links \textit{and}  missing ones, hence discarding the information in $\GraphTrain$ is wasteful.
This, in addition to the fact that %
purely-neural SoTA CQA methods might implicitly exploit this aspect at test time if they are able to memorize well $\GraphTrain$, pushes us to evaluate also \textit{hybrid} solvers.

{
Two remarkable examples are  QTO~\citep{QTO} and FIT~\citep{FIT}.
They explicitly retrieve existing links from $\GraphTrain$ when answering a query at test time. 
Inspired by them, and to evaluate our hypothesis, we create a light-weight hybrid solver, named \textbf{\textit{CQD-Hybrid}}, that is a variant of CQD that uses a pre-trained link predictor to compute scores for the answer entities of \ref{1p} queries only, denoting the unnormalized probability of that single link to exist \citep{NEURIPS2023_f4b76818}.
}
Then, assignments to existentially quantified variables in a complex query are greedily obtained by maximizing the combined score of the links, computed by replacing logical operators in the query with fuzzy ones \citep{van2022analyzing}.
A complete assignments to the variables (and hence to the target variable $T$), gives us an answer to the query.
In our CQD-Hybrid, we assign
the maximum score to those links that are observed in $\GraphTrain$.
That is, training links will have a higher score than the one that the link predictor would output, hence effectively steering the mentioned greedy procedure at test time.
This is the only minimal change we apply to CQD to test if its performance can depend on memorizing training triples. 
Note that this makes CQD-Hybrid different from QTO and FIT, which involve more sophisticated steps such as score calibration and a forward/backward update stage, that can boost performance. 
{In our experiment we will only consider QTO, as for the set of queries we considered QTO and FIT are equivalent \citep{FIT}.}
We report hyperparameters and implementation details of CQD-Hybrid in \cref{app:hyperparams_old}.

\textbf{All SoTA methods perform significantly worse on full-inference queries.}
We re-evaluate the SoTA methods mentioned above and stratify their performances based on the type of partial- and full-inference queries.
Where available, we reused the pre-trained models for their evaluation.
Otherwise, we re-produced the SoTA results using the hyperparameters provided by the authors, listed in 
\cref{app:hyperparams_old}.

\cref{tab:baselines_old_bench} presents a portion of the results in terms of MRR for the testing QA pairs
in the existing benchmarks \fbnew and \nell, along with the different QA pairs they can be reduced to when observing the training KG, grouped by each query type.
There, the performance of all SoTA methods consistently drops for query answer pairs that have a higher number of missing links in their reasoning tree.
Furthermore,  \cref{tab:NELL_complex query answering complex?per_answer,tab:fullcomparison?oldFB15k237} compare the MRR computed on all the available queries originally used in the benchmarks (column ``all''), and the scores on those query types that they can be reduced to. 
\textit{There is a high similarity in MRR between the columns ``all'' and ``1p'' that is evidence that the good performances of CQA methods are explained by their link prediction performances}, as a very high percentage of queries in fact are reduced to \ref{1p} queries (see \cref{tab:old-query-answers-type-stats}).
A similar conclusion can be drawn for queries including negation, as shown in \cref{tab:neg_old_MRR}.

Note that this happens consistently also for hybrid solvers such as QTO and our CQD-Hybrid: they have a harder time solving QA pairs with more than one missing link.
While their performance was consistently surpassing purely-neural solvers, and CQD-Hybrid consistently boosted the performance of CQD, their poor performance on harder QA pairs is not reflected in the overall average score that is usually reported on papers, see \cref{tab:baselines_old_bench}.

\begin{table}[!t]
\center
\setlength{\tabcolsep}{5pt}
\caption{\textbf{Exploiting training links during inference boosts MRR} on existing benchmarks, as shown by our CQD-Hybrid model compared to previous SoTA, for several query types. CQD-Hybrid always achieves higher MRR scores when compared to CQD, and outperforms more sophisticated non-hybrid baselines 9/14 times. Best values in bold and second best underlined. %
}
\vspace{3pt}
\scalebox{0.75}{
{
\begin{tabular}{crccccccc}
\toprule
& Model & 2p & 3p & 2i & 3i & 1p2i & 2i1p & 2u1p \\
\midrule
\multirow{6}{*}{\rotatebox{90}{\scalebox{.9}{\fbnew}}}
&GNN-QE & \underline{14.7} & \textbf{11.8} & \textbf{38.3} & \textbf{54.1} & \underline{31.1} & 18.9 & 9.7 \\
&ULTRAQ & 11.5 & 8.9 & 35.7 & 51.0 & 29.6 & 18.6 & 7.3 \\
&CQD & 13.2 & 7.8 & 35.0 & 48.5 & 27.5 & \underline{20.7} & \underline{10.5} \\
&ConE & 12.8 & 11.0 & 32.6 & 47.3 & 25.5 & 14.0 & 7.4 \\
&CLMPT & 13.9 & \underline{11.3} & 37.5 & 51.9 & 28.4 & 19.3 & \textbf{11.4} \\
&\textbf{CQD-Hybrid} & \textbf{15.0} & 11.0 & \underline{37.6} & \underline{52.7} & \textbf{31.2} & \textbf{24.0} & 10.3 \\
\midrule
\multirow{6}{*}{\rotatebox{90}{\scalebox{.9}{\nell}}}
&GNN-QE & 17.9 & 15.2 & 40.0 & 50.9 & 29.1 & 20.5 & 8.8 \\
&ULTRAQ & 11.2 & 9.7 & 36.3 & 47.7 & 25.1 & 15.6 & 8.4 \\
&CQD & 22.0 & 13.4 & \underline{42.2} & \underline{51.8} & \underline{31.5} & \underline{25.8} & 16.0 \\
&ConE & 16.0 & 13.8 & 39.6 & 50.2 & 26.1 & 17.6 & 11.1 \\
&CLMPT & \underline{22.1} & \textbf{18.2} & 41.6 & 51.7 & 29.0 & 24.7 & \underline{16.1} \\
&\textbf{CQD-Hybrid} & \textbf{23.8} & \underline{17.8} & \textbf{44.2} & \textbf{57.8} & \textbf{33.2} & \textbf{28.4} & \textbf{16.6} \\
\bottomrule
\end{tabular}
}}
\label{tab:CqdhybridvsSoTA}
\end{table}

\begin{figure}[!t]
    \centering
    \caption{\textbf{MRR of SoTA on the new benchmarks is significantly lower than the old ones } as reported for \fbnew and \nell (in blue) versus our new \fbnewH and \nellH (in orange)
    for different query types and SoTA models.
   }
    \includegraphics[width=\columnwidth]{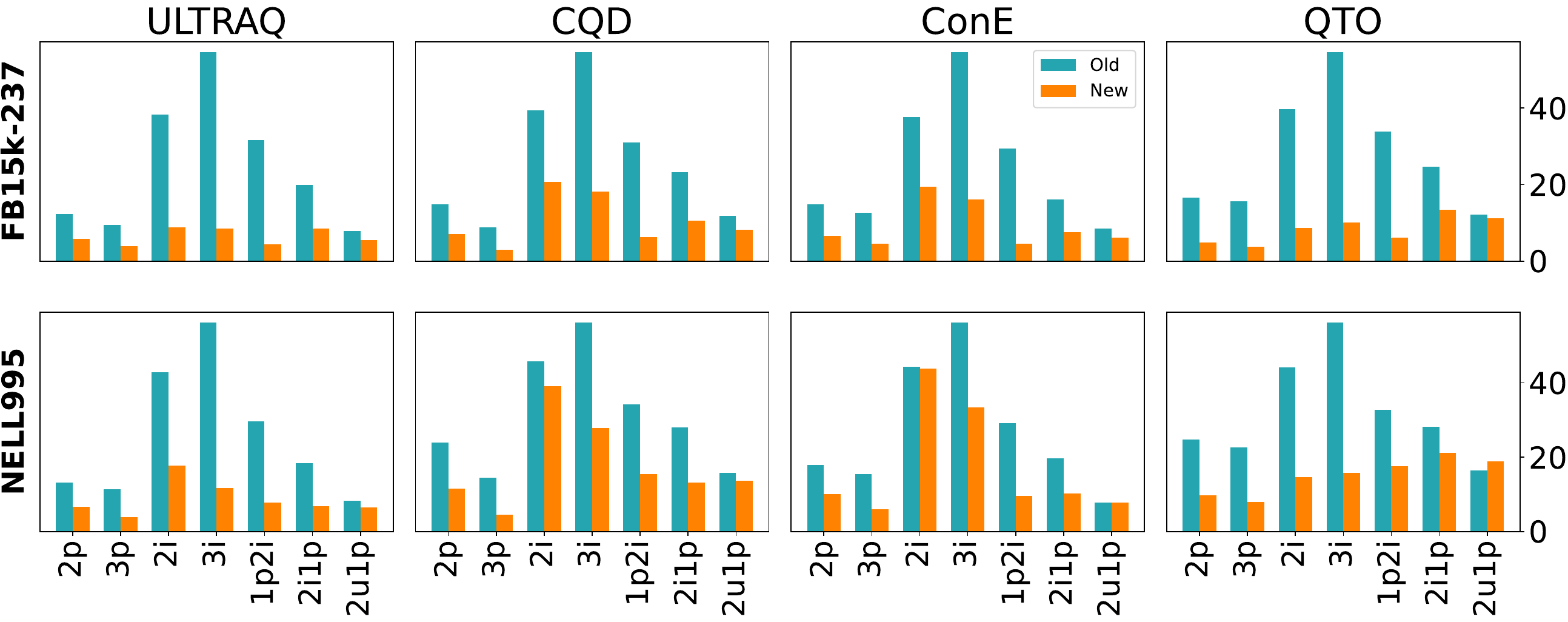}
    \label{fig:oldvsnewbenchs}
\end{figure}

\cref{tab:CqdhybridvsSoTA} highlights that our  CQD-hybrid is able to surpass existing non-hybrid SoTA, which is evidence that a pre-trained neural link predition plus memorization of $\GraphTrain$ achieves SoTA performances for \fbnew and \nell.
The complete stratified comparison for all solvers is reported in \cref{tab:NELL_complex query answering complex?per_answer,tab:fullcomparison?oldFB15k237}.
However, in a few cases it happens that full-inference results are higher than some partial-inference; in such cases, the reason is to be found in the fact that full-inference QA pairs are very scarce (see \cref{tab:old-query-answers-type-stats}).\footnote{We rule out that this is due to the influence of the number of existing entities bounding to existential quantified variables.
For example we report such an analysis for CQD in \cref{fig:MRR_card_old_bench,fig:prop_card_new_benchs}. Moreover, we found that no anchor entities nor relation names is predominant in both benchmarks, as shown in \cref{tab:imbalances?oldbenchs}.} \textbf{\textit{This motivates us to create a fully-balanced benchmark}} in the next section.
\begin{table}[!t]

\begin{minipage}{\linewidth}
\caption{\textbf{Queries of type 2u are less hard than previously thought}
in terms of MRR, with our filtered \ref{2u} queries having comparable performance to \ref{1p}, as discussed in \cref{sec:background}, in contrast with the results of the \ref{2u} queries originally used (\ref{2u} all).
}
\label{tab:2u_old_benc}
    \centering
    \scalebox{0.75}{
    \begin{tabular}{rccccccc}
    \toprule%
     \multicolumn{3}{r}{\fbnew} & \multicolumn{4}{r}{\nell}\\
     \cmidrule(lr){2-4}\cmidrule(lr){5-8}
     method & 1p&2u all & 2u filtered&& 1p& 2u all  & 2u filtered\\

     \midrule
\multirow{4}{*}{} %
  GNN-QE&42.8& 16.2
  &\textbf{40.7}&&53.6&15.4
  &34.8\\
ULTRAQ&40.6&13.2
&33.6 &&38.9&10.2
&21.3\\
  CQD &\textbf{46.7}&\textbf{17.6}  
  &32.7&&\textbf{60.4}&\textbf{19.9}
  &35.9\\
  ConE &41.8&14.9 
  &29.9&&60.0&14.9 
  &28.2\\
  CLMPT &45.3&14.8  
  &32.8&&58.1&18.6
  &\textbf{39.6}\\

  \bottomrule
   
\end{tabular}
}
\end{minipage}%
\end{table}
\par
\textbf{Non-existing links for union queries give a false sense of hardness.} 
For union queries, we filter query answer pairs, as discussed in \cref{sec:inv_exst_benchs}, removing those with non-existing links in their reasoning trees. 
In \cref{tab:2u_old_benc}, we show that if we do not do that, and consider also the pairs with non-existing links (denoted as ``2u all''),  MRR performance greatly drops. 
In this way, we reproduce the low performance for \ref{2u} queries that the original SoTA baselines reported in their papers.
However, as \cref{tab:2u_old_benc} shows, this is just an artifact due to including non-filtered pairs while computing \cref{eq_metrics}.
With our filtered pairs, instead, we recover the performance of \ref{1p} queries, as expected (\cref{sec:background}).
Similar conclusions can be drawn for other union queries as reported in \cref{tab:union_old_benc}.

\begin{takeawaybox}[label={takeaway:performances}]

\textbf{Takeaway 2.}
    \fbnew and \nell are not suitable to precisely assess  CQA methods, as most of their QA pairs can be predicted by just predicting a single missing links. This 
    results in highly inflated performance that distorts the perception of progress in the field and pushes the community to improve the performance only on the easiest QA pairs.
\end{takeawaybox}

%% file: new-benchmarks.tex
In this section, we aim to answer the following question: (\textbf{RQ3}) \emph{How can we construct a set of CQA benchmarks that let us measure the performance on each query type while not distorting the aggregated average?}
To do so, we present a new set of CQA benchmarks based on the previously used \fbnew and \nell, as well as a novel one we design out of temporal knowledge graph \icews to question the current way to split the original KG $G$ into $\GraphTrain$ and $\GraphTest$ (see our discussion of how sampling triples independently is arbitrary in \cref{sec:inv_exst_benchs}).
Then, we evaluate current SoTA methods on these new benchmarks to establish a rigorous baseline for future works. 
\par
\textbf{Building new CQA benchmarks.}
We generate our benchmarks to comprise 
both full-inference and partial-inference QA pairs, but we modify the algorithm of \cite{DBLP:conf/iclr/RenHL20} to ensure that no training links are present in the reasoning trees of full-inference pairs, and for each query type, we 
{ build the benchmark such that the full spectrum of hardness appears with the same frequency in it.} 
For example, for \ref{3p} queries, we sample 30,000 QA pairs,  10,000 for \ref{3p} that can be reduced to \ref{1p}, 10,000 for those reduceable to \ref{2p}, and 10,000 for full-inference QA pairs (\ref{3p} QA pairs non-reduceable to any other type).
{
Furthermore, we sample union queries \ref{2u}, \ref{2u1p} such that non-existing link are not present in their reasoning tree.}
Finally, to raise the bar of ``complexity'' in CQA, we introduce two query types that, in their full-inference versions, are harder than simpler types by design.
Specifically, we design ``4p'' queries for a sequential path made of four links and ``4i'', which represents the intersection of the target entity sets defined over four links. For their logical formulation, see \cref{4p4iform}.

We do so for the query types shown in \cref{fig:query_types,fig:query_types_neg}, and name the \textit{harder} versions of \fbnew and \nell, {\fbnewH} and {\nellH}. Only exceptions are \ref{1p}, \ref{2in}, \ref{2nu1p} those only being of type full-inference. See \cref{app:an_neg,app:negation_queries} for details about queries including negation, %
and \cref{app:new_bench_stats} for details about query generation.

\textbf{Non-uniform at random splits.}
Recall from \cref{sec:inv_exst_benchs} that \fbnew and \nell have been created by assuming triple independence and by inheriting the train/test split from their link prediction counterparts.
For \fbnewH and \nellH, we removed the triple independence assumption as we sample the same number of QA pairs per type, but we keep the existing data splits of $\GraphTrain$ and $\GraphTest$, for retro-compatibility. 
However, to evaluate the impact of this artificial splitting process, we adopt a more realistic one for \icewsH, where $\GraphTrain$ contains only ``past'' links, and we might want to predict the future links contained in $\GraphTest$.
To this end, we leverage the temporal information in \icews by (1) ordering the links based on their timestamp; (2) removing the temporal information, thus obtaining regular triples; and (3) selecting the train set to be the first temporally-ordered 80\% of triples, the valid the next 10\%, and the remaining to be the test split.
If the same fact appears with multiple timestamps, we retain only the link with the earliest timestamp.
For detailed statistics about the splits, please refer to \cref{app:kg_statistics}.
By doing so, we create a much more challenging benchmark even for easy \ref{1p} queries.

\par
\textbf{How do SoTA methods fare on our new benchmarks?}
For \fbnewH and \nellH we re-used the pre-trained models used for the experiments in \cref{sec:inv_exst_benchs}, while for the newly created \icewsH we trained GNN-QE, CLMPT, ConE, and the link predictor used in CQD, CQD-Hybrid and QTO from scratch. %
As this is not necessary for ULTRAQ, being a zero-shot neural link prediction applicable to any KG, we re-used the checkpoint provided by the authors.
Hyperparameters of all models are in  \cref{app:hyperparams_new}.

\cref{fig:oldvsnewbenchs} reports the MRR on the new benchmarks versus the old benchmarks, where the MRR of \fbnew and \nell (in blue) is much higher than the one of \fbnewH and \nellH (in orange) for all methods for all query types.
Considering how MRR can drop for more than 30 points, e.g., for 3i queries, this highlights once more how distorted the perception of progress in the field can be.

\cref{tab:new-bench+h-MRR} shows the results of the selected baselines on the new benchmarks, for all query types in \cref{fig:query_types,fig:query_types_neg}. 
For stratified results on \fbnewH, \nellH, \icewsH see \cref{tab:strat-FB-bal,tab:strat-NELL-bal,tab:strat-ICEWS18-bal}, respectively. For the stratified analysis on queries including negation see \cref{tab:neg_prop_MRR}.
\textbf{\textit{The surprising result from \cref{tab:new-bench+h-MRR} is that no model obtains SoTA performance on most query type}}, which was not the case for the old benchmarks, where QTO obtained SoTA performance for most query types (see \cref{tab:fullcomparison?oldFB15k237,tab:NELL_complex query answering complex?per_answer,tab:neg_old_MRR}). 
This is further evidence that current SoTA methods mainly focused on improving the performance on the easiest QA pairs and likely overfit the benchmarks.
Instead, with our new benchmarks, this will not be possible in the same way, as we compile the benchmark with the same amount of partial- and full-inference QA pairs per type.

Furthermore, we highlight how full-inference \ref{2u} queries have similar performance with \ref{1p}, similar to how full-inference \ref{2u1p} are expected to be as hard as \ref{2p}, as discussed in \cref{sec:hardness}. 
Finally, we remark that \icewsH is much harder than \nellH and \fbnewH, across all query types, even \ref{1p}, thus raising the bar for neural link predictors as well. 
This confirms our hypothesis that the sampling method of the KG splits plays a big role in determining the hardness of the benchmark.
{For additional analysis on the new benchmarks, comprising the influence of the number of existing entities bounding to existential quantified variables, the most frequent relation name and anchor entity per query type, and statistical tests, see \cref{app:analysis_new_bench}.}

\begin{table*}[!t]
\center
\setlength{\tabcolsep}{5.5pt}
\caption{ \textbf{There is no clear SoTA method for the new benchmarks.} As shown in \cref{fig:oldvsnewbenchs}, the MRR on the new benchmarks is significantly lower than the old ones. For example, for 3i queries on \fbnewH, QTO has an MRR of 10.1, while for \fbnew it was 54.6. For the stratified analysis see \cref{tab:strat-FB-bal,tab:strat-NELL-bal,tab:strat-ICEWS18-bal,tab:neg_prop_MRR}. 
}
\scalebox{.88}{
\vspace{3pt}
\begin{tabular}{crcccccccccccccccc}
\toprule
& Model & 1p & 2p & 3p & 2i & 3i & 1p2i & 2i1p &2u& 2u1p & 4p& 4i& 2in& 3in& 2pi1pn& 2nu1p& 2in1p\\
\midrule
\multirow{6}{*}{\rotatebox{90}{\scalebox{.9}{\fbnewH}}}  &
  GNN-QE & 42.8 & 5.2 & 4.0 & 6.0 & 8.8 & 5.6 & 9.9 & 32.5 & 10.0 & 4.3 & 20.0&6.8&\textbf{6.5}&\textbf{3.7}&\textbf{5.0}&3.3\\ 
  & ULTRAQ& 40.6 & 4.5 & 3.5 & 5.2 & 7.2 & 5.3 & 10.1 & 29.4 & 8.3 & 3.8 & 16.4&5.3&5.5&2.6&3.7&2.2\\ 
  &CQD & \textbf{46.7} & 4.4 & 2.4 & \textbf{11.3} & \textbf{12.8} & 6.0 & 11.5 & 40.1 & 10.6 & 1.1 & 23.8&3.3&2.6&0.6&4.2&1.2\\ 
  &CQD-Hybrid & \textbf{46.7} & 4.8 & 3.1 & 6.0 & 8.6 & 5.5 & 12.9 & \textbf{42.2} & 12.0 & 2.4 & 17.4&4.7&1.6&1.0&2.4&1.3\\ 
  &ConE & 41.8 & 4.6 & 3.9 & 9.1 & 10.3 & 3.8 & 7.9 & 22.8 & 6.0 & 3.5 & 20.3&5.1&4.9&2.9&3.3&\textbf{3.6}\\ 
  &QTO & \textbf{46.7} & 4.9 & 3.7 & 8.7 & 10.1 & \textbf{6.1} & 13.5 & 30.6 & 11.2 & 3.9 & 20.2&\textbf{10.6}&3.1&2.0&3.3&1.5\\ 
  &CLMPT & 45.3 & \textbf{5.3} & \textbf{4.7} & 10.2 & 12.2 & 5.6 & \textbf{14.9} & 33.6 & \textbf{14.2} & \textbf{4.5} & \textbf{24.0} &5.4&2.3&1.6&3.8&2.5\\ 
\midrule
\multirow{6}{*}{\rotatebox{90}{\scalebox{.9}{\nellH}}}
&GNN-QE& 53.6 & 8.0 & 6.0 & 10.7 & 13.3 & 16.0 & 13.5 & 47.5 & 9.8 & 4.7 & 19.4
& 5.5 & \textbf{6.4} & \textbf{5.8} & 3.3 & 4.4\\ 
&ULTRAQ& 38.9 & 6.1 & 4.1 & 7.9 & 10.2 & 15.8 & 9.3 & 28.1 & 9.5 & 4.2 & 15.6
& 4.5 & 5.9 & 4.3 & 2.7 & 3.6\\ 
&CQD& \textbf{60.4} & 9.6 & 4.2 & 18.5 & 19.6 & 18.9 & 22.6 & 46.3 & 18.5 & 2.0 & 24.8
& 4.2 & 1.5 & 1.5 & 3.6 & 2.6\\ 
&CQD-Hybrid& \textbf{60.4} & 9.0 & 6.1 & 12.1 & 14.4 & 17.4 & 21.2 & 46.4 & 19.3 & 3.5 & 20.4
& 5.1 & 1.2 & 1.4 & 2.9 & 2.4\\ 
&ConE& 53.1 & 7.9 & 6.7 & 21.8 & 23.6 & 14.9 & 11.8 & 39.9 & 8.8 & 5.2 & 27.6& 4.6 & 6.0 & 3.7 & 2.7 & \textbf{6.4}\\ 
&QTO & 60.3 & 9.8 & \textbf{8.0} & 14.6 & 15.8 & \textbf{17.6} & 21.1 & 49.1 & 18.9 & 7.0 & 20.9
& \textbf{11.3} & 2.3 & 3.1 & \textbf{4.3} & 2.4\\ 
&CLMPT & 58.1 & \textbf{10.1} & 7.8 & \textbf{22.7} & \textbf{25.0} & 17.2 & \textbf{24.4} & \textbf{50.0} & \textbf{22.0} & \textbf{7.2} & \textbf{29.1}& 5.2 & 2.4 & 2.2 & 3.1 & 4.0\\ 
\midrule
\multirow{6}{*}{\rotatebox{90}{\scalebox{.9}{\icewsH}}}
&GNN-QE & 12.2 & 0.9 & 0.5 & \textbf{16.1} & \textbf{26.5} & \textbf{19.1} & 3.5 & \textbf{7.6} & 1.1 & 0.4 & \textbf{34.0}
& 4.5 & 6.9 & 0.9 & \textbf{3.5} & 0.8\\  
&ULTRAQ & 6.3 & 1.2 & 1.2 & 7.0 & 11.7 & 8.8 & 1.3 & 3.3 & 1.2 & 0.8 & 15.9
& 2.3 & 4.8 & \textbf{1.2} & 2.2 & \textbf{1.6}\\  
&CQD & \textbf{16.6} & 2.5 & \textbf{1.5} & 13.0 & 19.5 & 17.1 & \textbf{6.7} & 6.8 & \textbf{5.9} & \textbf{1.1} & 24.0
& 1.5 & 2.9 & 0.2 & 2.7 & 0.9\\  
&CQD-Hybrid& \textbf{16.6} & \textbf{2.6} & \textbf{1.5} & 15.0 & 25.5 & 17.5 & 5.8 & 6.8 & 5.6 & 0.9 & 33.2
& 1.7 & 4.0 & 0.3 & 2.2 & 1.1\\
&ConE & 3.5 & 0.9 & 0.9 & 1.2 &0.5 & 1.2 & 1.6 & 1.1 & 0.9 & 0.6 & 0.3
& 1.7 & 2.9 & 1.1 & 0.9 & 1.3\\  
&QTO & \textbf{16.6} & \textbf{2.6} & 1.4 & 15.7 & 25.0 & 18.4 & 6.2 & 6.7 & 4.9 & \textbf{1.1} & 31.5
& \textbf{4.9} & \textbf{8.7} & \textbf{1.2} & 3.0 & 0.9\\  
&CLMPT & 4.7 & 0.8 & 0.1 & 12.0 & 23.0 & 9.7 & 2.1 & 2.7 & 2.2 & 0.1 & 31.0& 1.2 & 2.1 & 0.1 & 1.0 & 0.2\\ 
  \bottomrule
\end{tabular}
}
\label{tab:new-bench+h-MRR}
\end{table*}

\begin{takeawaybox}[label={takeaway:complex-benchmarks}]

\textbf{Takeaway 3.}
    {Our benchmarks are built with the same }amount
    of partial-inference and full-inference QA pairs that depend on the query type, allowing to measure CQA performance while not distorting the aggregated performance across all QA pairs. Additionally, \icewsH highlights that more realistic sampling strategies are challenging for current SoTA. 
\end{takeawaybox}

%% file: appendix.tex
\appendix
\onecolumn
\counterwithin{table}{section}
\counterwithin{figure}{section}
\renewcommand{\thetable}{\thesection.\arabic{table}}
\renewcommand{\thefigure}{\thesection.\arabic{figure}}

\newpage
\section{Additional analysis on the old benchmarks}\label{app:analysis_old_bench}
\subsection{Influence of existing links on queries involving negation} \label{app:an_neg}

In \cref{tab:neg-analysis} we show an analysis for queries involving negation. For such analysis, we split the reasoning tree, defined in \cref{sec:hardness}, into two subparts, namely \textit{positive reasoning tree}, composed by the non-negated triples, and the \textit{negative reasoning tree}, only composed by the negated triples. In particular, in the same spirit of \cref{tab:old-query-answers-type-stats}, by only looking at the positive reasoning tree, we report the percentage of QA pairs that can be reduced to easier types (partial-inference) and the one that cannot be reduced to a simpler type (full-inference). Furthermore, we argue that also the negative reasoning tree contains training triples, but how this propagates to performance is less clear than the positive part, as each method treats negation differently. \cref{tab:neg-analysis} shows that in both \fbnew and \nell, queries can be reduced to easier types, revealing that our analysis also extends to negated queries. The only exceptions are \ref{2in}, \ref{2nu1p} where the positive reasoning tree consists of a single link, see \cref{app:negation_queries}. Consequently, no reduction to partial inference is possible (see \cref{sec:hardness}).

\begin{table}[ht!]
\centering
\small
\caption{
\textbf{Most negated queries have training links in the positive reasoning tree of the QA pairs}. Partial-inference and full-inference refer to the \textbf{positive} reasoning tree. '-' refers to reductions that are not possible.
}
\begin{tabular}{rrrrrr}
\toprule
& \multicolumn{2}{c}{\fbnew} & \multicolumn{3}{c}{\nell} \\
\cmidrule(lr){2-3} \cmidrule(lr){5-6}
& partial-inference& full-inference && partial-inference& full-inference\\
\midrule
2in  & -&100&& -&100\\
3in  & 95.4&4.6&& 93.9&6.1\\
2pi1pn  & 98.4&1.6&& 97.8&2.2\\
2nu1p  & -&100&& -&100\\
2in1p  & 97.4&2.6&& 95.6&4.4\\
\bottomrule
\end{tabular}
\label{tab:neg-analysis}
\end{table}

\begin{table}[ht!]
\caption{\textbf{Even for queries including negation the MRR results are highly influenced by the presence of existing links in the positive reasoning tree of the QA pairs.} For example, for 3in queries of \fbnew, the overall MRR of QTO is 16.1, while its MRR on the portion of full-inference QA pairs is only 2.1. Hence, the overall result is highly influence by the vast number of partial inference QA pairs in the benchmark (see \cref{tab:neg-analysis})
}
\small
    \centering
    \begin{tabular}{crccccccc}
    \toprule%
     \multicolumn{4}{r}{\fbnew} & \multicolumn{4}{r}{\nell}\\
     \cmidrule(lr){3-5}\cmidrule(lr){7-9}
     Query type & method & overall 
& partial-inf & full-inf & & overall 
& partial-inf & full-inf \\

     \midrule
\multirow{4}{*}{2in} %
  & GNN-QE& 6.8&-&6.8&&5.5&-&5.5\\
& ULTRAQ& 5.3&-&5.3&&4.5&-&4.5\\
& CQD& 3.3&-&3.3&&4.2&-&4.2\\
& CQD-hybrid& 4.7&-&4.7&&5.1&-&5.1\\
& QTO& \textbf{10.6}&-&\textbf{10.6}&&\textbf{11.3}&-&\textbf{11.3}\\
& ConE& 5.1&-&5.1&&4.6&-&4.6\\
& CLMPT& 5.4&-&5.4&&5.2&-&5.2\\
  \midrule
  \multirow{4}{*}{3in} %
  & GNN-QE& 11.3&11.7&1.1&&9.0&9.5&0.6\\
& ULTRAQ& 9.6&9.9&0.6&&8.2&8.6&0.7\\
& CQD& 7.9&8.0&\textbf{2.7}&&6.5&6.7&1.1\\
& CQD-hybrid& 10.4&10.7&1.1&&8.3&8.8&0.7\\
& QTO& \textbf{16.1}&16.7&2.1&&\textbf{14.2}&14.9&1.2\\
& ConE& 8.0&8.3&1.7&&6.5&6.8&\textbf{1.3}\\
& CLMPT& 10.8&11.1&1.7&&6.6&6.9&1.2\\
  \midrule
  \multirow{4}{*}{2pi1pn} %
  & GNN-QE& 4.8&4.8&1.8&&3.7&3.7&1.5\\
& ULTRAQ& 3.6&3.6&1.5&&3.2&3.2&1.7\\
& CQD& 2.0&2.0&0.6&&2.2&2.2&1.4\\
& CQD-hybrid& 3.1&3.1&1.1&&3.1&3.2&0.8\\
& QTO& \textbf{8.9}&9.0&\textbf{2.5}&&\textbf{7.4}&7.5&\textbf{2.7}\\
& ConE& 4.1&4.1&1.7&&3.0&3.1&1.6\\
& CLMPT& 4.0&4.0&1.8&&3.2&3.2&1.7\\
  \midrule
  \multirow{4}{*}{2nu1p} %
  & GNN-QE& \textbf{5.0}&-&\textbf{5.0}&&3.3&-&3.3\\
& ULTRAQ& 3.7&-&3.7&&2.7&-&2.7\\
& CQD& 4.2&-&4.2&&3.6&-&3.6\\
& CQD-hybrid& 2.4&-&2.4&&2.9&-&2.9\\
& QTO& 3.3&-&3.3&&\textbf{4.3}&-&\textbf{4.3}\\
& ConE& 3.3&-&3.3&&2.7&-&2.7\\
& CLMPT& 3.8&-&3.8&&3.1&-&3.1\\
  \midrule
  \multirow{4}{*}{2in1p} %
  & GNN-QE& 5.2&5.3&1.7&&5.4&5.5&2.7\\
& ULTRAQ& 3.8&3.8&1.4&&4.7&4.8&1.9\\
& CQD& 3.6&3.6&1.4&&5.5&5.5&2.6\\
& CQD-hybrid& 4.1&4.1&1.5&&6.6&6.7&2.5\\
& QTO& \textbf{7.9}&8.1&1.6&&7.4&7.6&2.0\\
& ConE& 6.3&6.4&\textbf{3.0}&&7.5&7.6&4.0\\
& CLMPT& 6.1&6.1&\textbf{3.0}&&\textbf{7.7}&7.8&\textbf{4.2}\\
  \bottomrule
   
\end{tabular}
    \label{tab:neg_old_MRR}
\end{table}

\subsection{Additional comparisons}
\cref{tab:fullcomparison?oldFB15k237} and \cref{tab:NELL_complex query answering complex?per_answer} show the full results of \fbnew and \nell, including the results comparing all methods, all query types, and the subtypes they are reduced to. %

Moreover, \ref{tab:union_old_benc} show the full results of the union queries, including the old overall results, which include QA pairs having non-existing links in their reasoning tree (see \cref{fig:2u-query-reduction}), and the ``new overall'', in which those QA pairs are filtered out. In particular, for ``2u1p'' queries, we observe that for \fbnew, the full-inference queries have a higher MRR than the overall, while for \nell, we find the MRR results on full-inference QA pairs to be lower than the new overall results as expected. We attribute the results on \fbnew to the presence of bias in the QA pairs due to their scarcity in the benchmarks. We argue that for the same reason, 2u1p performances are not comparable to 2p. However, we will show that for the full-inference QA pairs in the new benchmark, our claim is confirmed (see \cref{tab:strat-FB-bal,tab:strat-NELL-bal,tab:strat-ICEWS18-bal}).

Moreover, for query types including negation, results in \cref{tab:neg_old_MRR} show the same pattern of \cref{tab:baselines_old_bench}, i.e. the MRR of the full-inference QA pairs is much lower than the one of the partial-inference QA pairs and of the overall.

\begin{table}[ht]
\caption{\textbf{Full inference \ref{2u} QA pairs show higher performance than non-filtered, while full-inference \ref{2u1p} pairs show lower or comparable performance.} Results on union queries for existing benchmarks and comparison between old and new overalls. `-' refers to reductions that are not possible, while `/' to reductions that are possible but that are not present in the data. 
}
\small
    \centering
    \scalebox{0.9}{
    \begin{tabular}{crccccccccccc}
    \toprule%
     \multicolumn{5}{r}{\fbnew} & \multicolumn{6}{r}{\nell}\\
     \cmidrule(lr){3-7}\cmidrule(lr){8-13}
     Query type & method & overall 
& overall (new) & 1p & 2u&2u1p& & overall 
& overall (new) & 1p & 2u&2u1p \\

     \midrule
\multirow{4}{*}{2u} %
  & GNN-QE& \underline{16.2}
  &\textbf{40.7} & /&\textbf{40.7}& -&&15.4
  &34.8&/&34.8&-\\
& ULTRAQ& 13.2
&\underline{33.6} & /& \underline{33.6}& -&&10.2
&21.3&/&21.3&-\\
  & CQD &\textbf{17.6}  
  &32.7 & / & 32.7& -&&19.9
  &35.9&/&35.9&- \\
  & ConE & 14.3
  &29.9 & / &  29.9&-& &14.9 
  &28.2 &/&28.2 &- \\
  \midrule
  \multirow{4}{*}{2u1p}
& GNN-QE & \textbf{13.4} 
&9.7 & 8.7 & 50.9&13.1&&8.8
&8.8&6.3&53.1&2.0\\
& ULTRAQ& 10.2
&7.3 & 6.6 & 33.2&\textbf{15.0}&&8.4
&7.0&6.6&37.1&6.5\\
  & CQD&11.3 
  &10.5 & 10.4 & 13.7&\underline{14.7}&&16.0
  &14.6&14.2&51.8&9.5\\
  & ConE & 10.6 
  & 7.4 & 7.0 & 20.3& 11.9 && 11.1 
  & 7.0 &9.4&43.2&4.0  \\
  \bottomrule
   
\end{tabular}
   }
    \label{tab:union_old_benc}
\end{table}
\begin{table}[ht]
\centering
\caption{\textbf{The best model on all available \nell queries (column ``overall'') and the best on the full inference queries (diagonal) is different for 6/7 query types}, excluding 1p and 2u, being already of type full-inference. Comparison of MRR scores for SoTA methods for the different subtypes \nell queries can be reduced to. Best in bold. %
}
\scalebox{.73}{
\begin{tabular}{crcccccccccc}
\toprule
Query type & method &overall& 1p & 2p & 3p & 2i & 3i & 1p2i & 2i1p &2u& 2u1p\\
\midrule
1p 
& GNN-QE& 53.6 & 53.6 & - & -& -& -& -& -& -& -\\
& ULTRA-query& 38.9 & 38.9 & - & -& -& -& -& -& -& -\\
& ConE& 60.0 & 60.0 & - & -& -& -& -& -& -& -\\
& CQD& \textbf{60.4} & \textbf{60.4} & - & -& -& -& -& -& -& -\\
  & CQD-hybrid& \textbf{60.4} & \textbf{60.4} & - & -& -& -& -& -& -& -\\
    & QTO& 60.3 & 60.3 & - & -& -& -& -& -& -& -\\
    & CLMPT& 58.1 & 58.1 & - & -& -& -& -& -& -& -\\
  
\midrule
2p 
& GNN-QE& 17.9 & 18.2 & 6.1 & -& -& -& -& -& -& -\\
& ULTRA-query& 11.2 & 11.5 & 4.6 & -& -& -& -& -& -& -\\
& ConE& 16.0 & 16.3 & 7.2 & -& -& -& -& -& -& -\\
& CQD& 22.0 & 22.3 & 7.6 & -& -& -& -& -& -& -\\
& CQD-hybrid& 23.8 & 24.2 & 6.2 & -& -& -& -& -& -& -\\
& QTO& \textbf{24.7} & 25.1 & 7.3 & -& -& -& -& -& -& -\\
& CLMPT& 22.1 & 22.3 & \textbf{8.9} & -& -& -& -& -& -& -\\
  
\midrule
3p 
& GNN-QE& 15.2 & 15.1 & 8.1 & 3.4& -& -& -& -& -& -\\
& ULTRA-query& 9.7 & 9.8 & 5.1 & 4.1& -& -& -& -& -& -\\
& ConE & 13.8 & 13.8 & 8.2 & 6.5 & -& -& -& -& -& -\\
& CQD& 13.4 & 12.8 & 7.8 & \textbf{8.5}& -& -& -& -& -& -\\
  & CQD-hybrid& 17.8 & 17.2 & 9.9 & 8.1& -& -& -& -& -& -\\
    & QTO& \textbf{22.6} & 22.3 & 10.9 & 7.7& -& -& -& -& -& -\\
    & CLMPT& 18.3 & 17.7 & 10.7 & 8.2& -& -& -& -& -& -\\
  
\midrule
2i 
& GNN-QE& 40.0 & 41.4 & - & -& 3.6& -& -& -& -& -\\
& ULTRA-query& 36.3 & 37.5 & - & -& 3.2& -& -& -& -& -\\
& ConE& 39.6 & 40.6 & - & -& 8.0 & -& -& -& -& -\\
& CQD& 42.2 & 43.3 & - & -& 6.9& -& -& -& -& -\\
& CQD-hybrid& \textbf{44.2} & 45.7 & -& - & 6.7& -& -& -& -& -\\
& QTO& \textbf{44.2} & 45.7 & - & -& 5.1& -& -& -& -& -\\
& CLMPT& 41.6 & 42.5 & - & -& \textbf{8.9}& -& -& -& -& -\\
  
\midrule
3i 
& GNN-QE& 50.9 & 53.6 & - & -& 10.9& 2.2& -& -& -& -\\
& ULTRA-query& 47.7 & 50.0 & - & -& 11.3& 1.5& -& -& -& -\\
& ConE & 50.2 & 51.9 & - & -& 18.3 & 4.8 & -& -& -& -\\
& CQD& 51.8 & 53.9 & - &  -&14.4& 3.6& -& -& -& -\\
  & CQD-hybrid& \textbf{57.8} & 61.6 & - &  -&10.7& 2.5& -& -& -& -\\
    & QTO& 56.2 & 60.0 & - & -& 12.4& 2.4& -& -& -& -\\
    & CLMPT& 51.7 & 53.4 & - & -& 20.3& \textbf{5.3}& -& -& -& -\\
  
\midrule
1p2i 
& GNN-QE& 29.1 & 42.7 & 21.4 & -& 12.0& -& 10.2& -& -& -\\
& ULTRA-query& 25.1 & 39.0 & 25.6 & -&  8.2&-& 7.0& -& -& -\\
& ConE& 26.1 & 38.5 & 25.2 & -& 10.5 & -& 8.6 & -& -& -\\
& CQD& 31.5 & 44.0 & 25.7 & -& 14.5& -& \textbf{12.6}& -& -& -\\
& CQD-hybrid& \textbf{33.2} & 48.7 & 24.2 & -& 13.7& -& 9.4&-& -& -\\
& QTO& 32.8 & 47.4 & 22.0 & -& 14.5& -& 9.8& -& -& -\\
& CLMPT& 29.0 & 39.3 & 15.3 & -& 21.8& -& 12.1& -& -& -\\
  
\midrule
2i1p 
& GNN-QE& 20.5 & 20.8 & 16.3 & -& 14.6& -& -& 20.7& -& -\\
& ULTRA-query& 15.6 & 16.5 & 9.5 & -& 9.6& -& -& 11.4& -& -\\
& ConE& 17.6 & 17.7 & 16.4 & -& 25.7 & -& -& 19.5 & -& -\\
& CQD& 25.8 & 25.9 & 21.2 & -& 26.3& -& -& 23.6& -& -\\
  & CQD-hybrid& \textbf{28.4} & 28.9 & 20.4 & -& 22.8& -& -& 23.3& -& -\\
    & QTO& 28.2 & 28.5 & 20.2 & -& 24.0& -& -& 24.4& -& -\\
& CLMPT& 24.7 & 24.6 & 26.5 & -& 26.7& -& -& \textbf{25.2}& -& -\\    
  
\midrule
2u 
& GNN-QE& 34.8 & - & - & -& -& -& -& -& 34.8& -\\
& ULTRA-query& 21.3 & - & - & -& -& -& -& -& 21.3& -\\
& ConE& 28.2%
& - & - & -& -& -& -& -& 28.2 & -\\
& CQD& 35.9 & - & - & -& -& -& -& -&35.9 & -\\
& CQD-hybrid& 35.9 & - & - & -& -& -& -& -& 35.9& -\\
& QTO& 37.6 & - & - & -& -& -& -& -& 37.6& -\\
& CLMPT& \textbf{39.6} & - & - & -& -& -& -& -& \textbf{39.6}& -\\
  
\midrule
2u1p 
& GNN-QE& 8.8 & 6.3 & - & -& -& -& -& -& 53.1& 2.0\\
& ULTRA-query& 8.4 & 6.6 & - & -& -& -& -& -& 37.1& 
6.5\\
& ConE& 7.0 & 9.4 & - & -& -& -& -& -& 43.2 & 4.0 \\
& CQD& 14.6 & 14.2 & - & -& -& -& -& -&51.8 & 9.5\\
  & CQD-hybrid& 15.0 & 14.5 & - & -& -& -& -& -& 53.5& 9.6\\
  & QTO& 16.4 & 15.8 & - & -& -& -& -& -& 61.2& 9.6\\
  & CLMPT& \textbf{17.7} & 17.4 & - & -& -& -& -& -& 37.5& \textbf{10.2}\\

\bottomrule
\end{tabular}
}
\label{tab:NELL_complex query answering complex?per_answer}

\end{table}

\begin{table}[ht]
\centering
\caption{\textbf{The best model on all available \fbnew queries (column ``overall'') and the best on the full inference queries (diagonal) is different for 5/7 query types}, excluding 1p and 2u, those (q,a) pairs being all of type full-inference.  Comparison of MRR scores for SoTA methods for the different subtypes \fbnew queries can be reduced to. Best in bold.  %
}
\label{tab:fullcomparison?oldFB15k237}
\scalebox{.73}{
\begin{tabular}{crcccccccccc}
\toprule
Query type & method &overall& 1p & 2p & 3p & 2i & 3i & 1p2i & 2i1p & 2u & 2u1p\\
\midrule
1p 
& GNN-QE& 42.8 & 42.8 & - & -& -& -& -& -& -& -\\
& ULTRA-query& 40.6 & 40.6 & - & -& -& -& -& -& -& -\\
 & ConE& 41.8 & 41.8 & - & -& -& -& -& -& -& -\\
& CQD& 46.7 & 46.7 & - & -& -& -& -& -& -& -\\
  & CQD-hybrid& 46.7 & 46.7 & - & -& -& -& -& -& -& -\\
  & QTO& \textbf{46.7} & \textbf{46.7} & - & -& -& -& -& -& -& -\\
  & CLMPT& 45.3 & 45.3 & - & -& -& -& -& -& -& -\\
\midrule
2p 
  & GNN-QE& 14.7 & 14.8 & 4.7 & -& -& -& -& -& -& -\\
  & ULTRA-query& 11.5 & 11.5 & 4.2 & -& -& -& -& -& -& -\\
  & ConE& 12.8 & 12.8 & \textbf{5.2} & -& -& -& -& -& -& -\\
  & CQD& 13.2 & 13.3 & 3.5 & -& -& -& -& -& -& -\\
  & CQD-hybrid& 15.0 & 15.2 & 3.5 & -& -& -& -& -& -& -\\
    & QTO& \textbf{16.6} & 16.7 & 4.0 & -&-& -& -& -& -& -\\
    & CLMPT& 13.9 & 13.9 & 5.1 & -&-& -& -& -& -& -\\
\midrule
3p 
& GNN-QE& 11.8 & 12.0 & 4.4 & 4.8 & -& -& -& -& -& -\\
  & ULTRA-query& 8.9 & 9.0 & 4.6 & 4.4& -& -& -& -& -& -\\
  & ConE & 11.0 & 11.0 & 5.4 & 2.6 & -& -& -& -& -& -\\
  & CQD& 7.8 & 7.8 & 3.4 & 1.8& -& -& -& -& -& -\\
  & CQD-hybrid& 11.0 & 11.1 & 3.6 & 4.4& -& -& -& -& -& -\\
    & QTO& \textbf{15.6} & 15.8 & 4.5 & \textbf{5.0}& -& -& -& -& -& -\\
    & CLMPT& 11.3 & 11.3 & 6.1 & 4.0& -& -& -& -& -& -\\
\midrule
2i 
& GNN-QE& 38.3 & 39.3 & - & -& 4.0& -& -& -& -& -\\
  & ULTRA-query& 35.7 & 36.7 & - & -& 3.4& -& -& -& -& -\\
  & ConE& 32.6 & 33.3 & - & -& 5.5 & -& -& -& -& -\\
  & CQD& 35.0 & 35.8 & - & -& \textbf{7.3} & -& -& -& -& -\\
  & CQD-hybrid& 37.6 & 38.7 & -& - & 3.8& -& -& -& -& -\\
    & QTO& \textbf{39.7} & 40.8 & - & -& 5.7& -& -& -& -& -\\
    & CLMPT& 37.5 & 38.4 & - & -& \textbf{7.3}& -& -& -& -& -\\
\midrule
3i 
& GNN-QE& 54.1 & 56.0 & - & -& 10.9& 5.2& -& -& -& -\\
  & ULTRA-query& 51.0 & 52.9 & - & -& 9.7& 4.3& -& -& -& -\\
  & ConE& 47.3 & 48.3 & - & -& 15.6 & 4.0 & -& -& -& -\\
  & CQD& 48.5 & 49.6 & - &  -&17.0& \textbf{6.0} & -& -& -& -\\
  & CQD-hybrid& 52.7 & 54.8 & - &  -&9.9& 4.6& -& -& -& -\\
    & QTO& \textbf{54.6} & 56.4 & - & -& 15.4& 5.4& -& -& -& -\\
    & CLMPT& 51.9 & 53.1 & - & -& 17.7& 5.9& -& -& -& -\\
\midrule
1p2i 
& GNN-QE& 31.1 & 32.8 & 15.5 & -& 5.7& -& 9.1 & -& -& -\\
  & ULTRA-query& 29.6 & 31.5 & 19.2 & -&  4.2&-& 7.4 & -& -& -\\
  & ConE & 25.5 & 26.6 & 13.9 & -& 5.4 & -& \textbf{9.7} & -& -& -\\
  & CQD& 27.5 & 28.7 & 13.4 & -& 7.1& -& 9.0& -& -& -\\
  & CQD-hybrid& 31.2 & 33.4 & 14.8 & -&  4.8& -& 7.0&-& -& -\\
    & QTO& \textbf{33.8} & 35.9 & 15.8 & -& 6.2& -& 7.3& -& -& -\\
    & CLMPT& 28.4 & 30.0 & 16.2 & -& 5.9& -& 8.3& -& -& -\\
\midrule
2i1p 
& GNN-QE& 18.9 & 19.2 & 8.2 & -& 6.2& -& -& 3.4& -& -\\
  & ULTRA-query& 18.6 & 18.9 & 8.5 & -& 5.1& -& -& 8.1 & -& -\\
  & ConE& 14.0 & 13.8 & 9.7 & -& 12.8 & -& -& 5.6 & -& -\\
  & CQD& 20.7 & 21.0 & 10.5 & -& 6.7& -& -& 7.6& -& -\\
  & CQD-hybrid& 24.0 & 24.6 & 10.0 & -& 4.6& -& -& 7.5& -& -\\
    & QTO& \textbf{24.7} & 25.3 & 10.3 & -& 8.6& -& -& 8.1& -& -\\
    & CLMPT& 19.3 & 19.1 & 18.7 & -& 7.9& -& -& \textbf{13.2}& -& -\\
\midrule
2u 
& GNN-QE& \textbf{40.7} & - & - & -& -& -& -& -&\textbf{40.7}& -\\
  & ULTRA-query& 33.6 & - & - & -& -& -& -& -&33.6& -\\
  & ConE& 29.9 & - & - & -& -& -& -& -& 29.9& -\\
  & CQD& 32.7 & - & - & -& -& -& -& -&32.7& -\\
  & CQD-hybrid& 32.7 & - & - & -& -& -& -& -& 32.7& -\\
& QTO& 37.0 & - & - & -& -& -& -& -& 37.0& -\\
& CLMPT& 32.8 & - & - & -& -& -& -& -& 32.8& -\\
\midrule
2u1p 
& GNN-QE& 9.7 & 8.7 & - & -& -& -& -& -& 50.9& 13.1\\
  & ULTRA-query& 7.3 & 6.6 & - & -& -& -& -& -& 33.2& 15.0\\
  & ConE& 7.4 & 7.0 & - & -& -& -& -& -& 20.3& 11.9\\
  & CQD& 10.5 & 10.4 & - & -& -& -& -& -&13.7& 14.7\\
  & CQD-hybrid& 10.3 & 10.0 & - & -& -& -& -& -& 18.5& 13.6\\
    & QTO& 12.2 & 11.6 & - & -& -& -& -& -& 35.3& 11.3\\
    & CLMPT& \textbf{12.7} & 12.5 & - & -& -& -& -& -& 10.6& \textbf{27.7}\\
\bottomrule
\end{tabular}
}

\end{table}

\paragraph{Performance analysis of QTO}
The stratified MRR performance of QTO \citep{QTO} for the old benchmarks have been included in \cref{tab:fullcomparison?oldFB15k237,tab:NELL_complex query answering complex?per_answer}. For a fair comparison, we remark that we used the same link predictor for CQD, CQD-Hybrid, and QTO. Details and hyperameters are available in \cref{app:hyperparams_old}. Our analysis reveals that, similarly to the other baselines, QTO performance drops when evaluated on the full-inference QA pairs only (diagonal), w.r.t overall results. Moreover, even when QTO is the SoTA on a certain query type, most of the time it is not so on the portion of full-inference QA pairs only, showing that improving performance on the partial-inference QA pairs not necessarily results on improvements over the full-inference ones. 

Moreover, results on \nell, shown in \cref{tab:NELL_complex query answering complex?per_answer}, reveal that CQD-Hybrid outperforms QTO for some query types of \nell, i.e. \ref{3i}, \ref{1p2i}, \ref{2i1p}, suggesting that even by only setting a score=1 to the training and validation triples it is possible to obtain SoTA results on the old benchmarks. We remark that, while both in CQD-Hybrid and QTO a score=1 is set to the training and validation links, QTO is much more sophisticated than CQD-Hybrid, as they 1) calibrated the scores with a heuristics, 2) store a matrix $|V|\times|V|$ for each relation name containing the score for every possible triples, 3) have a forward/backward mechanism in the reasoning.

\subsection{Influence of intermediate existing entities on the results}
For query types having intermediate variables, i.e., \ref{2p}, \ref{3p}, \ref{2i1p}, \ref{1p2i}, and \ref{2u1p}, we analyzed how the number of existing intermediate entities matching the existentially quantified variables could influence the results.
We refer to this number as \textit{cardinality of existing entities}.
For example, in \cref{fig:2i1p-query-reduction} the entities ``When in Rome" and ``Spiderman 2" are existing intermediate entities matching the existential quantified variable $V$ for the query \ref{query2i1pexample}.
The intuition is that while some of these entities simplify the prediction of the answers of the query (see \cref{sec:hardness}), some others might be misleading for the prediction. %
Hence, if there is a high number of existing entities in the path, it would be harder for the model to select only the existing entities needed to predict the next link. For this reason, we claim that a high value of cardinality of existing entities would make a QA pair \textit{harder}.

\begin{figure}[ht]
    \centering
    \begin{minipage}{1\textwidth}
        \centering
        \caption{\textbf{Higher cardinality of intermediate existing entities leads to lower MRR.} Influence of the number of existing intermediate variables on the MRR for datasets \fbnew and \nell, using CQD.}
        \label{fig:MRR_card_old_bench}
        \includegraphics[width=0.9\linewidth]{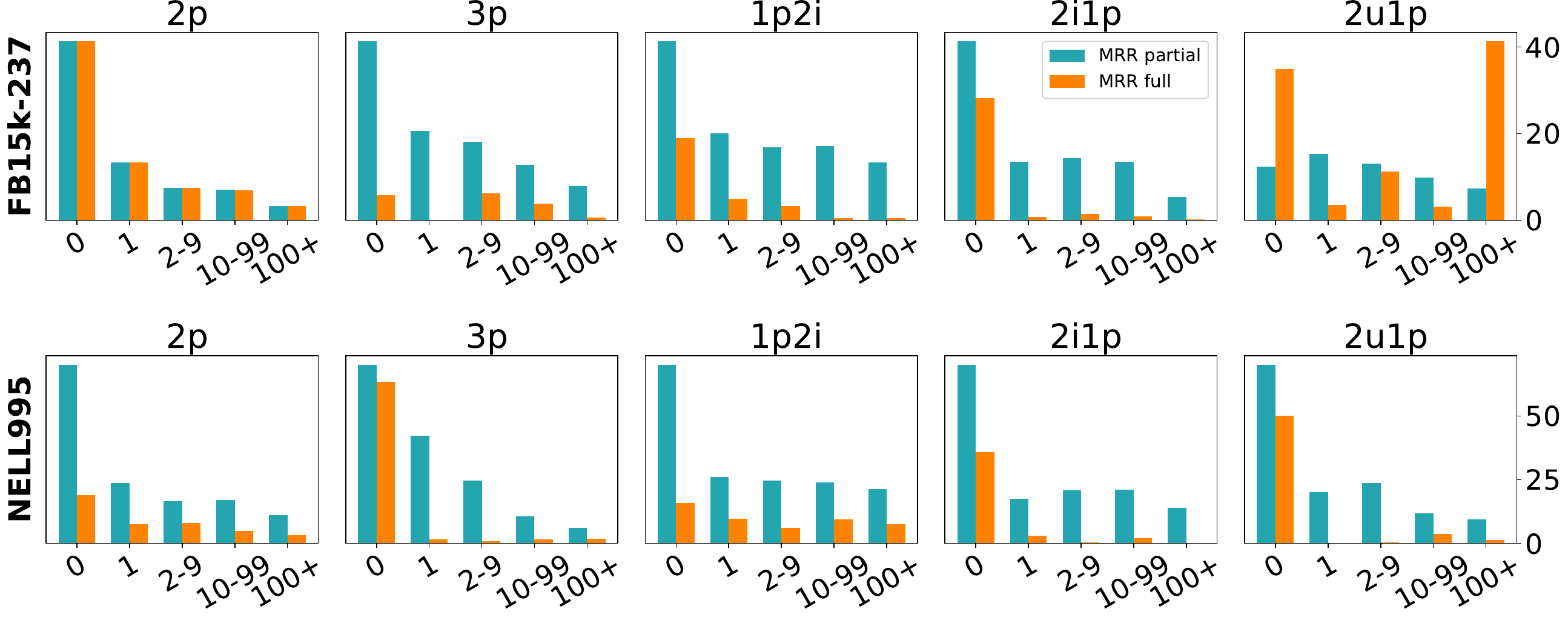}
    \end{minipage}
    \vspace{0.2cm}
    \begin{minipage}{1\textwidth}
        \centering
        \caption{\textbf{Proportion ofQA pairs having different cardinality of intermediate existing entities in the old benchmarks.}}
        \label{fig:prop_card_old_bench}
        \includegraphics[width=0.9\linewidth]{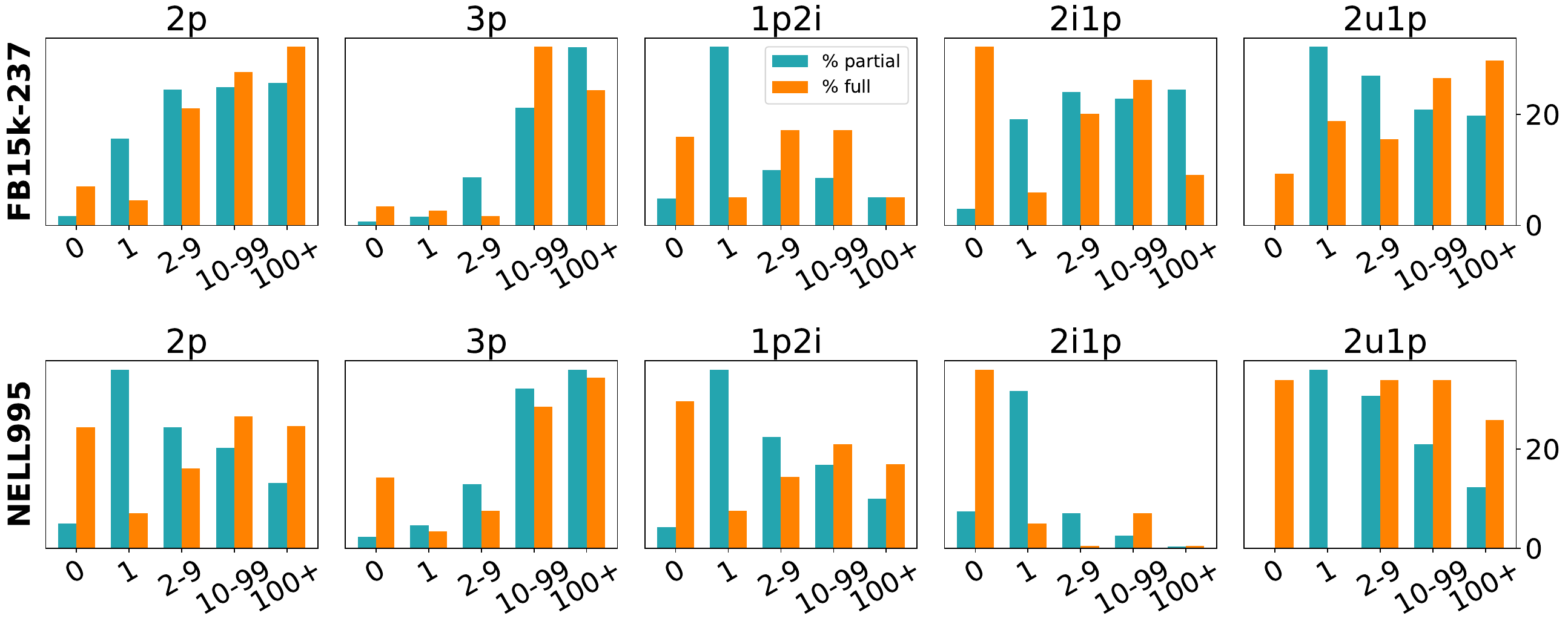}
    \end{minipage}
\end{figure}

To support our claim, we analyze the percentage of QA pairs having different values of cardinality, %
shown in 
\cref{fig:prop_card_old_bench} %
, and their MRR, using CQD \citep{DBLP:conf/iclr/ArakelyanDMC21}, shown in \cref{fig:MRR_card_old_bench}. 
\cref{fig:MRR_card_old_bench} shows that CQD results are highly influenced by the value of cardinality, having decreasing performance at the increase of the cardinality. Note that by matching this plot with the one in \cref{fig:prop_card_old_bench}, one can understand if results on a specific query type are highly influenced by a high proportion of QA pairs having the same cardinality. For example, the MRR of \ref{1p2i} partial-inference QA pairs in \fbnew is highly influenced by the presence of about 30\% of QA pairs having cardinality of ``1''.

\begin{table}[t]
\centering
\caption{\textbf{There is no significant unbalancement in anchor nor relation names for each query type }. Most frequent relation name and anchor per query type in the old benchmarks. Highest value in bold. %
}
\label{tab:imbalances?oldbenchs}
\scalebox{0.9}{
\begin{tabular}{lcccc}
\toprule

Query & relation name & frequency(\%) & anchor entity & frequency(\%) \\
\midrule
\multicolumn{5}{c}{\fbnew} \\
\midrule
1p & %
film\_release\_region
& 7.5 & USA & 0.7 \\
2p & location/location/contains & 8.6& USA%
& 4.0 \\
3p & location/location/contains & 11.6 & USA & 3.3 \\
2i & people/person/profession & 16.4 & USA & 16.4 \\
3i & people/person/gender & 18.7& USA & \textbf{30.1} \\
pi & people/person/profession & 13.7 & USA & 14.1 \\
ip & people/person/profession & 8.6 & US dollars%
& 5.7 \\
2u & film/film\_crew\_role & 14.6 & USA & 22.5 \\
2u1p & %
taxonomy\_entry/taxonomy 
& 9.9 & /m/08mbj5d & 5.4 \\
2in & people/person/profession & 15.7 & USA & 15.2\\
3in & common/webpage/category & \textbf{20.7} & USA & 28.5\\
2pi1pn & people/person/profession & 12.4 & USA & 11.7\\
2nu1p & people/person/profession & 17.1 & USA & 14.5\\
2in1p & location/contains & 13.4 & USA & 14.3\\

\midrule
\multicolumn{5}{c}{\nell} \\
\midrule
1p & concept:atdate & 9.4 & concept\_stateorprovince\_california & 0.5 \\
2p & concept:mutualproxyfor & 8.7 & concept\_lake\_new & 1.5 \\
3p & concept:mutualproxyfor & 15.9 & concept\_book\_new & 2.5 \\
2i & concept:atdate & 22.5& concept\_book\_new & 7.6 \\
3i & concept:atdate & 21.6 & concept\_company\_pbs & 12.0 \\
pi & concept:proxyfor & 11.4 & concept\_book\_new & 6.1 \\
ip & concept:subpartof & 9.4 & concept\_athlete\_sinorice\_moss & 3.8 \\
2u & concept:atdate & 10.8 & concept\_company\_pbs & 4.9 \\
2u1p & concept:subpartof & 11.1 & concept\_athlete\_sinorice\_moss & 3.3\\
2in & concept:atdate & 20.9 & concept\_book\_new & 6.8\\
3in & concept:atdate & \textbf{45.8} & concept\_book\_new & \textbf{14.2}\\
2pi1pn & concept:proxyfor & 14.9 & concept\_book\_new & 5.7\\
2nu1p & concept:proxyfor & 21.9 & concept\_book\_new & 8.0\\
2in1p & concept:atdate & 15.0 & concept\_lake\_new & 6.1\\
\bottomrule
\end{tabular}
}

\end{table}

\subsection{Imbalances of relation names and anchor nodes} Next, we check if there is some imbalances in both relation names and anchors. Note that if a relation name or an anchor entity is present multiple times in a query, this is counted only once. As shown in \cref{tab:imbalances?oldbenchs}, for \fbnew and \nell we find that in most cases there is no predominant relation name nor anchor entity. However, there are exceptions, as for example, for anchor entities of 3i queries of \fbnew, where the anchor node ``USA'' is present in 30.1\% of them, and for \ref{2u} queries, where it is present in 22.5\%. We also notice that there is a predominance of USA as an anchor entity across the vast majority of query types, which is most likely given by the vast presence of this entity in the knowledge graph.

\newpage
\clearpage
\section{Query types including negation}\label{app:negation_queries}

\begin{figure*}[!t]
    \centering
    \begin{tabular}{cccccc}
         \includegraphics[height=.06\textwidth,page=17]{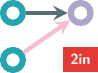}
         & 
         \includegraphics[height=.06\textwidth,page=18]{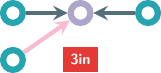}
         & 
         \includegraphics[height=.06\textwidth,page=19]{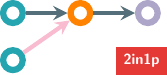}
         & 
         \includegraphics[height=.06\textwidth,page=20]{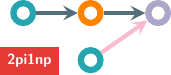}
         & 
         \includegraphics[height=.06\textwidth,page=21]{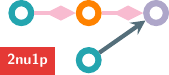}
         &
          \raisebox{-4pt}{\includegraphics[height=.08\textwidth,page=22]{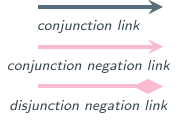}}
         \\
    \end{tabular}
    \caption{\textbf{Query structures including negation}, adapted from %
    \citet{DBLP:conf/nips/RenL20}, where ``2in1p'' queries are referred as ``inp''.
We explicitly mention the number of steps involved in a path or conjunction, as this is a factor of complexity. Analogously, ``pin'', ``pni'' queries from \citet{DBLP:conf/nips/RenL20} are now ``2pi1np'', and ``2nu1p''. See \cref{app:negation_queries} for their logical formulation.}
    \label{fig:query_types_neg}
\end{figure*}

In \cref{fig:query_types_neg} we include the query types including the negation operator involved in the analysis described in \cref{app:an_neg}. Those queries are obtained by adding a negation to some of the query structures described in \cref{sec:background}. For example, by negating a triple pattern involved in the intersection of a \ref{2i}, \ref{3i}, \ref{2i1p} queries, we obtain respectively:
\begin{align}
\label{2in}
\tag{2in} &?T:(\anchorA_1, \relR_1, \varTarget) \land \neg(\anchorA_2, \relR_2, \varTarget),\\
\tag{3in} &?T:(\anchorA_1, \relR_1, \varTarget) \land
    (\anchorA_2, \relR_2, \varTarget) \land
    \neg (\anchorA_3, \relR_3, \varTarget),\label{3in}\\
\tag{2in1p} &?T:\exists \varQuantified_1 .(
    (\anchorA_1, \relR_1, \varQuantified_1) \land
    \neg(\anchorA_2, \relR_2, \varQuantified_1) \land
    (\varQuantified_1, \relR_3, \varTarget)),\label{2in1p}
\end{align}
Moreover, by placing the negation in different triple patterns of the query type \ref{1p2i}, we obtain two query types:
\begin{align}
\label{2pi1pn}
\tag{2pi1pn} ?T:\exists \varQuantified_1 .(
    (\anchorA_1, \relR_1, \varQuantified_1) \land
    (\varQuantified_1, \relR_2, \varTarget) \land
    \neg(\anchorA_2, \relR_3, \varTarget)),
\end{align}
where the negation is placed in the triple pattern directly involved in the intersection, and
\begin{align}
 \tag{n2pi1p} ?T:\exists \varQuantified_1 .( \neg(
    (\anchorA_1, \relR_1, \varQuantified_1) \land
    (\varQuantified_1, \relR_2, \varTarget)) \land
    (\anchorA_2, \relR_3, \varTarget)),\label{n2pi1p}   
\end{align}
where the negation is placed on the path query \ref{2p}. 
Using De Morgan laws, this query can also be rewritten as:
\begin{align}
 \tag{2nu1p} ?T:\exists \varQuantified_1 .( 
    \neg(\anchorA_1, \relR_1, \varQuantified_1) \lor
    \neg(\varQuantified_1, \relR_2, \varTarget)) \land
    (\anchorA_2, \relR_3, \varTarget)),\label{2nu1p}   
\end{align}

Note that the visual representation of \ref{2nu1p} provided by \citep{DBLP:conf/nips/RenL20} was misleading, as in practice the negation is applied on the whole \ref{2p} query, rather than just on one of its links.

\subsection{Trivial answers of queries including negation}\label{app:easy-answers}
Recall from \cref{sec:hardness} that trivial $(q,t)$ pairs, i.e., pairs whose answer $t$ can be entirely retrieved from $\GraphTrain$, are filtered out from current CQA benchmarks~\citep{DBLP:conf/nips/RenL20}.
For queries without negation, every entity retrieved from $\GraphTrain$ is also an answer of $q$ on $G$, since $\GraphTrain \subseteq G$, while for queries including negation this does not hold.
For example, consider a query $q$ including a negated triple pattern, such as $\neg(\anchorA_2, \relR_2, \varTarget)$ in \ref{2in}.
If the link $(\anchorA_2, \relR_2, t)$ belongs to $\GraphTest$, the entity $t$ satisfies the negation on $\GraphTrain$, where this link is missing, but not on $G$.
Hence, $t$ can be retrieved from $\GraphTrain$, although it is not an answer of $q$ on $G$.
In the original \fbnew and \nell~\citep{DBLP:conf/nips/RenL20}, such entities are filtered out as trivial answers, hence they are never ranked as non-answers, which inflates the MRR.
We filter out only the entities retrieved from $\GraphTrain$ that are also answers of $q$ on $G$, for all query types of \fbnewH, \nellH, and \icewsH, including \ref{2in} and \ref{2nu1p}, which in \fbnewH and \nellH we keep from \citet{DBLP:conf/nips/RenL20}, and for the queries including negation of \fbnew and \nell.
The results in \cref{tab:new-bench+h-MRR,tab:neg_old_MRR,tab:neg_prop_MRR} are computed with these filters.

\section{More related works}
\label{app:more-related-works}
\citet{DBLP:conf/emnlp/GuML15} propose compositional training for embedding methods to predict answers for path queries.
GQE~\citep{DBLP:conf/nips/HamiltonBZJL18} learns a geometric intersection operator to answer conjunctive queries in embedding space; this approach was later extended by Query2Box~\citep{DBLP:conf/iclr/RenHL20}, BetaE~\citep{DBLP:conf/nips/RenL20}, and GNN-QE~\citep{DBLP:conf/icml/Zhu0Z022}.
FuzzQE~\citep{DBLP:conf/aaai/ChenHS22} improves embedding methods with t-norm fuzzy logic, which satisfies the axiomatic system of classical logic.
Some recent works such as HypE~\citep{DBLP:conf/www/ChoudharyRKSR21} and ConE~\citep{DBLP:conf/nips/ZhangWCJW21} use geometric interpretations of entity and relation embeddings to achieve desired properties for the logical operators.
Other solutions to CQA combine neural methods with symbolic algorithms.
For example, EmQL~\citep{DBLP:conf/nips/SunAB0C20} ensembles an embedding model and a count-min sketch, and is able to find logically entailed answers, while CQD~\citep{DBLP:conf/iclr/ArakelyanDMC21,DBLP:conf/nips/ArakelyanMDCA23} extends a pretrained knowledge graph embedding model to infer answers for complex queries.

\clearpage
\newpage
\section{Additional Analysis on the new benchmarks}\label{app:analysis_new_bench}
\subsection{Influence of intermediate existing entities on the results}
\begin{figure*}[!t]
    \centering
    \begin{minipage}{1\textwidth}
        \centering
        \includegraphics[width=0.8\linewidth]{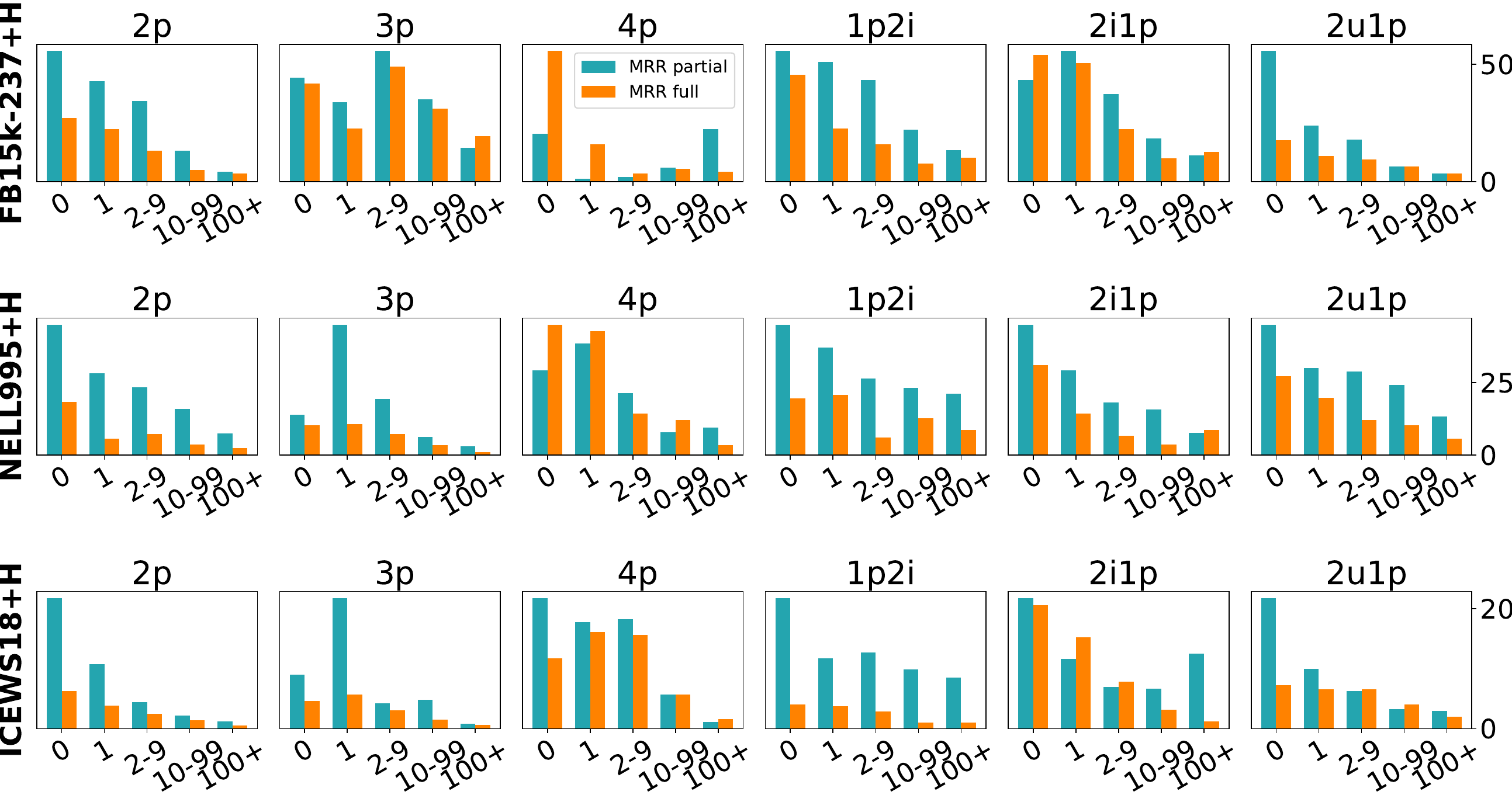}
        \caption{\textbf{Higher cardinality of intermediate existing entities leads to lower MRR.} Influence of the number of existing intermediate variables on the MRR for datasets \fbnewH, \nellH, \icewsH, using CQD.}
        \label{fig:MRR_card_new_benchs}
    \end{minipage}
    \vspace{0.5cm} %
    \begin{minipage}{1\textwidth}
        \centering
        \includegraphics[width=0.8\linewidth]{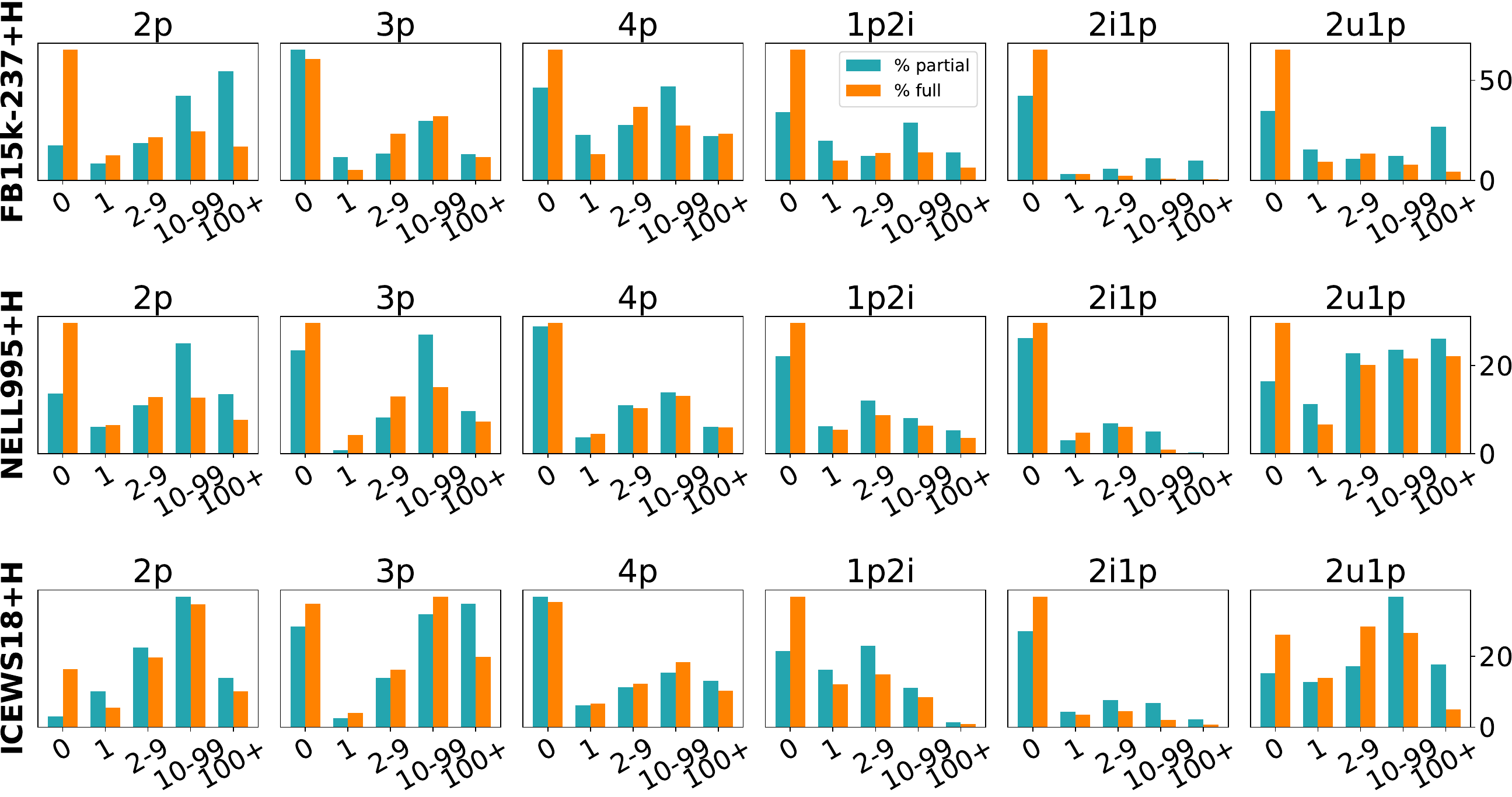}
        \caption{\textbf{Proportion ofQA pairs having different cardinality of intermediate existing entities in new benchmarks.}}
        \label{fig:prop_card_new_benchs}
    \end{minipage}
\end{figure*}

Similarly to what was done for the old benchmarks in \cref{app:analysis_old_bench}, we analyze the proportion and influence (on the MRR) of the number of intermediate existing entities for CQD, on the new benchmarks \fbnewH, \nellH, \icewsH. 
\cref{fig:MRR_card_new_benchs} show a very similar result to the one in \cref{fig:MRR_card_old_bench}, showing a decreasing MRR at the increase of the value of cardinality for all benchmarks. Moreover, for reference, we also report the proportion of the different QA pairs for each cardinality category in \cref{fig:prop_card_new_benchs}.

\begin{table}[ht]
\centering
\small
\scalebox{0.7}{
\begin{tabular}{llrlr}
\toprule

Query & relation name & frequency(\%) & anchor entity & frequency(\%) \\
\midrule
\multicolumn{5}{c}{\fbnewH} \\
\midrule
1p & film\_release\_region & 7.5 & USA & 0.7 \\
2p & /people/person/nationality& 15.3 & /m/08mbj5d/m/08mbj5d & 3.6 \\
3p & /people/person/profession& 14.2 & /m/092ys\_y & 5.4 \\
4p & /dataworld/gardening\_hint/split\_to& 20.0 & /m/0dq\_5 & 5.5 \\
2i & /people/person/profession& 17.2 & /m/09c7w0 & 15.5 \\
3i & /people/person/profession& 19.6 & /m/09c7w0 & 19.7 \\
4i & /people/person/profession& 19.2 & /m/09c7w0 & 20.0 \\
pi & /people/person/nationality& 19.7 & /m/09c7w0 & 14.9 \\
ip & /people/person/gender& 18.5 & /m/048\_p & 8.9 \\
2in & people/person/profession& 15.7 & USA & 15.2 \\
3in & /people/person/gender& 20.0 & /m/09c7w0 & 20.0 \\
2pi1pn & /people/person/nationality& 18.4 & /m/08mbj5d & 13.3 \\
2nu1p & people/person/profession & 17.1 & USA & 14.5 \\
2in1p & /people/person/profession& 13.6 & /m/08mbj5d & 12.2 \\
2u & /people/person/profession& 15.6 & /m/09c7w0 & 12.1 \\
2u1p & /common/topic/webpage./common/webpage/category& 19.4 & /m/0kc9f & 14.5 \\
\midrule
\multicolumn{5}{c}{\nellH} \\
\midrule
1p & concept:atdate& 9.4 & concept\_stateorprovince\_california & 0.5 \\
2p & concept:proxyfor& 17.9 & concept\_sportsteam\_ncaa\_mens\_midwest\_regionals & 6.4 \\
3p & concept:proxyfor& 20.0 & concept\_geopoliticallocation\_national & 5.9 \\
4p & concept:atlocation& 20.0 & concept\_website\_new\_york\_times & 6.5 \\
2i & concept:atdate& 20.0 & concept\_dateliteral\_n2006 & 5.4 \\
3i & concept:atdate& 20.0 & concept\_sportsleague\_mlb & 5.7 \\
4i & concept:atdate& 20.0 & concept\_date\_n2004 & 7.9 \\
pi & concept:proxyfor& 15.6 & concept\_book\_new & 7.8 \\
ip & concept:atdate& 19.4 & concept\_lake\_new & 9.4 \\
2in & concept:atdate& 20.9 & concept\_book\_new & 6.8 \\
3in & concept:atdate& 20.0 & concept\_sportsteamposition\_center & 15.5 \\
2pi1pn & concept:proxyfor& 20.0 & concept\_lake\_new & 15.2 \\
2nu1p & concept:proxyfor& 21.9 & concept\_book\_new & 8.0 \\
2in1p & concept:atdate& 20.0 & concept\_lake\_new & 11.2 \\
2u & concept:atdate& 20.0 & concept\_lake\_new & 2.7 \\
2u1p & concept:proxyfor& 19.8 & concept\_lake\_new & 11.0 \\
\midrule
\multicolumn{5}{c}{\icewsH} \\
\midrule
1p & Make statement& 18.8 & United States & 3.0 \\
2p & Make statement& 20.0 & Russia & 6.7 \\
3p & Consult& 20.0 & United States & 3.4 \\
4p & Make statement& 20.0 & Yukio Edano & 4.4 \\
2i & Make statement& 20.0 & Citizien(India) & 7.0 \\
3i & Make statement& 20.0 & Citizien(India) & 4.2 \\
4i & Make statement& 20.0 & United States & 4.5 \\
pi & Make statement& 20.0 & Unspecified Actor & 9.7 \\
ip & Make statement& 20.0 & Russia & 6.6 \\
2in & Make statement& 19.9 & Unspecified Actor & 17.1 \\
3in & Make statement& 20.0 & Citizien(India) & 20.0 \\
2pi1pn & Make statement& 20.0 & United States & 11.2 \\
2nu1p & Make statement& 19.9 & Unspecified Actor & 16.46 \\
2in1p & Consult& 20.0 & United States & 10.5 \\
2u & Consult& 19.9 & China & 3.5 \\
2u1p & Make statement& 19.9 & United States & 4.9 \\
\bottomrule
\end{tabular}
}
\caption{\textbf{The new benchmarks are generated such that anchors and relation names cannot appear more than 20\% in each query type.} Most frequent relation name and anchor per query type in the new benchmarks.} %
\label{tab:imbalances?new benchs}
\end{table}

\pagebreak

\paragraph{Imbalances of relation names and anchor nodes}
When creating the new benchmarks \fbnewH, \nellH, \icewsH, to make sure that no anchor entities nor relation name was predominant in each query type, we set a maximum of 20\% of QA pairs having the same anchor entity or relation name across all benchmarks and query type, as shown in \cref{tab:imbalances?new benchs}

\paragraph{Stratified analysis on new benchmarks}
For reference, we also provide the stratified analysis on the new benchmarks in \cref{tab:fullcomparison?oldFB15k237,tab:NELL_complex query answering complex?per_answer}, which shows that 1) the full-inference QA pairs have always a lower MRR than the overall, and 2) the overall MRR is much lower than the 1p-reductions only, as for the new benchmark the proportion of QA pairs is equally distributed.

The same considerations also apply to queries including negation, as shown in \cref{tab:neg_prop_MRR}.
{
\begin{table}[ht]
    \centering
    \scalebox{0.75}{
    \begin{tabular}{lccccccc}
    
        \toprule
        Model & CLMPT & CQD & CQD-h & ConE & GNNQE & QTO & ULTRA \\
        \midrule
        CLMPT  & --- & 3.32e-63 & 1.82e-80 & 3.92e-01 & 8.28e-71 & 1.06e-188 & 8.55e-35 \\
        CQD    & 3.32e-63 & --- & 1.48e-04 & 1.94e-52 & 1.72e-01 & 5.27e-38 & 2.31e-06 \\
        CQD-h  & 1.82e-80 & 1.48e-04 & --- & 8.88e-75 & 5.08e-03 & 8.41e-15 & 6.91e-16 \\
        ConE   & 3.92e-01 & 1.94e-52 & 8.88e-75 & --- & 6.65e-60 & 1.38e-165 & 7.04e-28 \\
        GNNQE  & 8.28e-71 & 1.72e-01 & 5.08e-03 & 6.65e-60 & --- & 4.79e-30 & 2.32e-09 \\
        QTO    & 1.06e-188 & 5.27e-38 & 8.41e-15 & 1.38e-165 & 4.79e-30 & --- & 2.22e-70 \\
        ULTRA  & 8.55e-35 & 2.31e-06 & 6.91e-16 & 7.04e-28 & 2.32e-09 & 2.22e-70 & --- \\
        \bottomrule
    \end{tabular}
    }
    \caption{Pairwise Mann–Whitney U test $p$-values on reciprocal ranks of 2p queries in \fbnewH. Lower values indicate stronger evidence of a significant difference between model performances.}
    \label{tab:stat-tests-2p}
\end{table}
}
\begin{table}[ht]
\centering
\caption{\textbf{Stratified analysis on \fbnewH}}
\scalebox{.63}{
\begin{tabular}{crcccccccccccc}
\toprule
Query type & method &overall& 1p & 2p & 3p & 4p & 2i & 3i & 4i & 1p2i & 2i1p &2u& 2u1p\\
\midrule
1p 
& GNN-QE& 42.8 & 42.8 & - & -& -& -& -& -& -& -& -& -\\
& ULTRA-query& 40.6 & 40.6 & - & -& -& -& -& -& -& -& -& -\\
& ConE& 41.8 & 41.8 & - & -& -& -& -& -& -& -& -& -\\
& CQD& 46.7 & 46.7 & - & -& -& -& -& -& -& -& -& -\\
  & CQD-hybrid& 46.7 & 46.7 & - & -& -& -& -& -& -& -& -& -\\
& QTO& 46.7 & 46.7 & - & -& -& -& -& -& -& -& -& -\\
& CLMPT& 45.3 & 45.3 & - & -& -& -& -& -& -& -& -& -\\
  
\midrule
2p 
& GNN-QE& 5.2 & 11.8 & 4.4 & -& -& -& -& -& -& -& -& -\\
& ULTRA-query& 4.5 & 9.5 & 3.9 & -& -& -& -& -& -& -& -& -\\
& ConE& 4.6 & 9.9 & 4.0 & -& -& -& -& -& -& -& -& -\\
& CQD& 4.5 & 7.4 & 4.1 & -& -& -& -& -& -& -& -& -\\
  & CQD-hybrid& 4.8 & 15.1 & 3.5 & -& -& -& -& -& -& -& -& -\\
& QTO& 4.9 & 14.4 & 3.8 & -& -& -& -& -& -& -& -& -\\
& CLMPT& 5.3 & 6.8 & 5.1 & -& -& -& -& -& -& -& -& -\\
  
\midrule
3p 
& GNN-QE& 4.0 & 16.2 & 4.6 & 3.6& -& -& -& -& -& -& -& -\\
& ULTRA-query& 3.5 & 14.0 & 3.3 & 3.2& -& -& -& -& -& -& -& -\\
& ConE& 3.9 & 10.1 & 5.0 & 3.6& -& -& -& -& -& -& -& -\\
& CQD& 2.4 & 3.7 & 2.2 & 2.4& -& -& -& -& -& -& -& -\\
  & CQD-hybrid& 3.1 & 10.7 & 3.8 & 2.7& -& -& -& -& -& -& -& -\\
& QTO& 3.7 & 21.4 & 5.8 & 3.0& -& -& -& -& -& -& -& -\\
& CLMPT& 4.7 & 8.3 & 4.1 & 4.5& -& -& -& -& -& -& -& -\\
\midrule
4p 
& GNN-QE& 4.3 & 16.1 & 14.1& 4.8&3.8 & -& -& -& -& -& -& -\\
& ULTRA-query& 3.8 & 14.4 & 12.3 & 4.6& 3.3& -& -& -& -& -& -& -\\
& ConE& 3.5 & 12.0 & 8.8 & 4.2& 3.2& -& -& -& -& -& -& -\\
& CQD& 1.1 & 0.3 & 0.2 & 0.7& 1.1& -& -& -& -& -& -& -\\
  & CQD-hybrid& 2.4 & 9.8 & 10.5 & 4.2& 2.0& -& -& -& -& -& -& -\\
& QTO& 3.9 & 18.2 & 21.7 & 5.5& 3.2& -& -& -& -& -& -& -\\ 
& CLMPT& 4.5 & 11.9 & 13.6 & 3.9& 4.1& -& -& -& -& -& -& -\\ 
\midrule
2i 
& GNN-QE& 6.0 & 20.4 & - & -& 5.4& -& -& -& -& -& -& -\\
& ULTRA-query& 5.2 & 20.7 & - & -& 4.6& -& -& -& -& -& -& -\\
& ConE& 9.1 & 15.5 & - & -& 8.8& -& -& -& -& -& -& -\\
& CQD& 11.3 & 18.4 & - & -& 10.8& -& -& -& -& -& -& -\\
  & CQD-hybrid& 6.0 & 22.0 & - & -& 5.4& -& -& -& -& -& -& -\\
& QTO& 8.7 & 24.4 & - & -& 8.0& -& -& -& -& -& -& -\\
& CLMPT& 10.2 & 18.4 & - & -& 9.7& -& -& -& -& -& -& -\\
  
\midrule
3i 
& GNN-QE& 8.8 & 32.2 & - & -& -& 2.5& 7.5& -& -& -& -& -\\
& ULTRA-query& 7.2 & 32.4 & - & -& -& 2.4& 5.9& -& -& -& -& -\\
& ConE& 10.3 & 25.0 & - & -& -& 6.9& 9.0& -& -& -& -& -\\
& CQD& 12.8 & 30.5 & - & -& -& 8.5& 11.5& -& -& -& -& -\\
  & CQD-hybrid& 8.6 & 38.2 & - & -& -& 2.7& 7.1& -& -& -& -& -\\
& QTO& 10.1 & 40.0 & - & -& -& 6.5& 8.2& -& -& -& -& -\\
& CLMPT& 12.2 & 31.8 & - & -& -& 7.6& 10.7& -& -& -& -& -\\

\midrule
4i
& GNN-QE& 19.6 & 47.9 & - & -& -& 18.3& 16.5& 15.8& -& -& -& -\\
& ULTRA-query& 16.4 & 48.9 & - & -& -& 17.5& 13.2& 12.9& -& -& -& -\\
& ConE& 20.3 & 36.9 & - & -& -& 24.9& 18.2& 15.8& -& -& -& -\\
& CQD& 23.8 & 45.1 & - & -& -& 27.0& 20.9& 19.8& -& -& -& -\\
  & CQD-hybrid& 17.4 & 58.3 & - & -& -& 17.2& 13.2& 14.6& -& -& -& -\\
& QTO& 20.2 & 60.2 & - & -& -& 25.1& 16.3& 15.6& -& -& -& -\\
& CLMPT& 24.0 & 40.5 & - & -& -& 26.7& 21.9& 18.7& -& -& -& -\\
\midrule

1p2i 
& GNN-QE& 5.6 & 28.1 & 6.3 & -& -& 3.0& -& -& 3.5& -& -& -\\
& ULTRA-query& 5.3 & 31.4 & 7.6 & -& -& 1.9& -& -& 3.1& -& -& -\\
& ConE& 3.8 & 16.7 & 5.1 & -& -& 2.5& -& -& 2.6& -& -& -\\
& CQD& 6.0 & 20.1 & 7.2 & -& -& 3.5& -& -& 4.2& -& -& -\\
  & CQD-hybrid& 5.5 & 33.2 & 7.7 & -& -& 2.6& -& -& 3.0& -& -& -\\
& QTO& 6.1 & 32.0 & 8.1 & -& -& 3.8& -& -& 3.6& -& -& -\\
& CLMPT& 5.6 & 18.1 & 5.5 & -& -& 26.5& -& -& 4.3& -& -& -\\
  
\midrule
2i1p 
& GNN-QE& 9.9 & 13.5 & 6.1 & -& -& 16.8& -& -& -& 7.2& -& -\\
& ULTRA-query& 10.1 & 13.8 & 5.7 & -& -& 18.4& -& -& -& 7.2& -& -\\
& ConE& 7.9 & 10.8 & 4.8 & -& -& 10.4& -& -& -& 7.2& -& -\\
& CQD& 11.5 & 12.2 & 6.4 & -& -& 21.7& -& -& -& 10.5& -& -\\
  & CQD-hybrid& 12.9 & 23.0 & 6.0 & -& -& 24.7& -& -& -& 8.2& -& -\\
& QTO& 13.5 & 23.5 & 6.6 & -& -& 27.3& -& -& -& 9.0& -& -\\
& CLMPT& 14.9 & 10.5 & 9.4 & -& -& 11.0& -& -& -& 14.2& -& -\\
  
\midrule
2u 
& GNN-QE& 32.4 & - & - & -& -& -& -& -& -& -& 32.4& -\\
& ULTRA-query& 29.4 & - & - & -& -& -& -& -& -& -& 29.4& -\\
& ConE& 22.8 & - & - & -& -& -& -& -& -& -& 22.8& -\\
& CQD& 40.0 & - & - & -& -& -& -& -& -& -& 40.0& -\\
  & CQD-hybrid& 42.2 & - & - & -& -& -& -& -& -& -& 42.2& -\\
& QTO& 30.6 & - & - & -& -& -& -& -& -& -& 30.6& -\\
& CLMPT& 33.6 & - & - & -& -& -& -& -& -& -& 33.6& -\\
  
\midrule
2u1p 
& GNN-QE& 10.0 & 7.6 & - & -& -& -& -& -& -& -& 54.0& 6.5\\
& ULTRA-query& 8.3 & 7.4 & - & -& -& -& -& -& -& -& 42.2& 4.8\\
& ConE& 6.0 & 6.0 & - & -& -& -& -& -& -& -& 17.1& 4.7\\
& CQD& 10.6 & 9.2 & - & -& -& -& -& -& -& -& 13.9& 8.4\\
  & CQD-hybrid& 12.0 & 10.5 & - & -& -& -& -& -& -& -& 44.0& 8.1\\
& QTO& 11.2 & 10.0 & - & -& -& -& -& -& -& -& 53.3& 6.7\\
& CLMPT& 14.2 & 8.3 & - & -& -& -& -& -& -& -& 9.1& 13.4\\

\bottomrule
\end{tabular}
}
\label{tab:strat-FB-bal}

\end{table}
\begin{table}[ht]
\centering
\caption{\textbf{Stratified analysis on \nellH}}
\scalebox{.63}{
\begin{tabular}{crcccccccccccc}
\toprule
Query type & method &overall& 1p & 2p & 3p & 4p & 2i & 3i & 4i & 1p2i & 2i1p &2u& 2u1p\\
\midrule
1p 
& GNN-QE& 53.6 & 53.6 & - & -& -& -& -& -& -& -& -& -\\
& ULTRA-query& 38.9 & 38.9 & - & -& -& -& -& -& -& -& -& -\\
& ConE& 60.0 & 60.0 & - & -& -& -& -& -& -& -& -& -\\
& CQD& 60.4 & 60.4 & - & -& -& -& -& -& -& -& -& -\\
  & CQD-hybrid& 60.4 & 60.4 & - & -& -& -& -& -& -& -& -& -\\
& QTO& 60.3 & 60.3 & - & -& -& -& -& -& -& -& -& -\\
& CLMPT& 58.1 & 58.1 & - & -& -& -& -& -& -& -& -& -\\
  
\midrule
2p 
& GNN-QE& 8.0 & 21.6 & 6.3 & -& -& -& -& -& -& -& -& -\\
& ULTRA-query& 6.1 & 16.2 & 5.0 & -& -& -& -& -& -& -& -& -\\
& ConE& 7.9 & 15.3 & 7.1 & -& -& -& -& -& -& -& -& -\\
& CQD& 9.6 & 21.7 & 8.0 & -& -& -& -& -& -& -& -& -\\
  & CQD-hybrid& 9.0 & 28.4 & 6.6 & -& -& -& -& -& -& -& -& -\\
& QTO& 9.8 & 25.9 & 7.7 & -& -& -& -& -& -& -& -& -\\
& CLMPT& 10.1 & 17.3 & 9.0 & -& -& -& -& -& -& -& -& -\\
  
\midrule
3p 
& GNN-QE& 6.0 & 33.3 & 14.1 & 3.9& -& -& -& -& -& -& -& -\\
& ULTRA-query& 4.1 & 14.7 & 9.4 & 3.1& -& -& -& -& -& -& -& -\\
& ConE& 6.7 & 21.0 & 12.2 & 5.5& -& -& -& -& -& -& -& -\\
& CQD& 4.2 & 6.2 & 5.6 & 3.7& -& -& -& -& -& -& -& -\\
  & CQD-hybrid& 6.1 & 22.2 & 10.9 & 4.4& -& -& -& -& -& -& -& -\\
& QTO& 8.0 & 42.1 & 15.3 & 5.2& -& -& -& -& -& -& -& -\\
& CLMPT& 7.8 & 20.1 & 10.6 & 6.4& -& -& -& -& -& -& -& -\\
\midrule
4p 
& GNN-QE& 4.7 & 4.9 & 25.0 & 11.9& 2.8& -& -& -& -& -& -& -\\
& ULTRA-query& 4.2 & 30.5 & 17.1 & 11.5& 2.9& -& -& -& -& -& -& -\\
& ConE& 5.2 & 27.3 & 17.3 & 7.5& 4.5& -& -& -& -& -& -& -\\
& CQD& 2.0 & 2.7 & 1.4 & 0.9& 2.0& -& -& -& -& -& -& -\\
  & CQD-hybrid& 3.5 & 34.9 & 19.3 & 8.1& 2.1& -& -& -& -& -& -& -\\
& QTO& 7.0 & 54.4 & 29.1 & 12.4& 4.9& -& -& -& -& -& -& -\\ 
& CLMPT& 7.2 & 30.0 & 18.7 & 7.6& 5.8& -& -& -& -& -& -& -\\ 
\midrule
2i 
& GNN-QE& 10.7 & 24.5 & - & -& 9.8& -& -& -& -& -& -& -\\
& ULTRA-query& 7.9 & 18.7 & - & -& 7.3& -& -& -& -& -& -& -\\
& ConE& 21.8 & 23.3 & - & -& 21.1& -& -& -& -& -& -& -\\
& CQD& 18.5 & 26.3 & - & -& 17.6& -& -& -& -& -& -& -\\
  & CQD-hybrid& 12.1 & 28.7 & - & -& 11.0& -& -& -& -& -& -& -\\
& QTO& 14.6 & 30.5 & - & -& 13.5& -& -& -& -& -& -& -\\
& CLMPT& 22.7 & 27.1 & - & -& 21.9& -& -& -& -& -& -& -\\
  
\midrule
3i 
& GNN-QE& 13.3 & 34.7 & - & -& -& 11.7& 10.6& -& -& -& -& -\\
& ULTRA-query& 10.2 & 26.7 & - & -& -& 11.4& 7.1& -& -& -& -& -\\
& ConE& 23.6 & 31.8 & - & -& -& 22.5& 21.5& -& -& -& -& -\\
& CQD& 19.6 & 34.6 & - & -& -& 17.4& 17.1& -& -& -& -& -\\
  & CQD-hybrid& 14.4 & 43.3 & - & -& -& 11.7& 11.8& -& -& -& -& -\\
& QTO& 15.8 & 46.2 & - & -& -& 14.1& 12.8& -& -& -& -& -\\
& CLMPT& 25.0 & 36.3 & - & -& -& 23.4& 22.2& -& -& -& -& -\\

\midrule
4i
& GNN-QE& 19.4 & 51.9 & - & -& -& 24.6& 14.3& 13.5& -& -& -& -\\
& ULTRA-query& 15.6 & 42.6 & - & -& -& 24.9& 12.2& 8.2& -& -& -& -\\
& ConE& 27.6 & 47.9 & - & -& -& 35.0& 21.8& 23.8& -& -& -& -\\
& CQD& 24.8 & 52.9 & - & -& -& 30.6& 17.7& 20.8& -& -& -& -\\
  & CQD-hybrid& 20.4 & 65.1 & - & -& -& 24.9& 14.2& 14.5& -& -& -& -\\
& QTO& 20.9 & 68.3 & - & -& -& 26.8& 13.8& 15.4& -& -& -& -\\
& CLMPT& 29.1 & 50.5 & - & -& -& 36.9& 22.4& 24.6& -& -& -& -\\
\midrule

1p2i 
& GNN-QE& 16.0 & 55.9 & 19.5 & -& -& 9.1& -& -& 7.3& -& -& -\\
& ULTRA-query& 15.8 & 55.0 & 23.0 & -& -& 5.5& -& -& 5.9& -& -& -\\
& ConE& 14.9 & 32.6 & 18.9 & -& -& 10.7& -& -& 7.0& -& -& -\\
& CQD& 18.9 & 54.3 & 22.1 & -& -& 12.0& -& -& 10.0& -& -& -\\
  & CQD-hybrid& 17.4 & 64.7 & 19.0 & -& -& 12.4& -& -& 8.1& -& -& -\\
& QTO& 17.6 & 58.4 & 20.6 & -& -& 13.4& -& -& 7.9& -& -& -\\
& CLMPT& 17.2 & 37.9 & 18.2 & -& -& 13.1& -& -& 10.6& -& -& -\\
  
\midrule
2i1p 
& GNN-QE& 13.5 & 26.2 & 12.9 & -& -& 17.0& -& -& -& 8.2& -& -\\
& ULTRA-query& 9.3 & 22.2 & 7.6 & -& -& 22.3& -& -& -& 6.0& -& -\\
& ConE& 11.8 & 22.6 & 11.1 & -& -& 23.2& -& -& -& 8.4& -& -\\
& CQD& 22.6 & 29.9 & 20.1 & -& -& 36.9& -& -& -& 16.1& -& -\\
  & CQD-hybrid& 21.2 & 35.8 & 18.9 & -& -& 34.5& -& -& -& 13.7& -& -\\
& QTO& 21.1 & 36.3 & 19.2 & -& -& 36.0& -& -& -& 13.4& -& -\\
& CLMPT& 24.4 & 25.9 & 21.8 & -& -& 35.2& -& -& -& 20.0& -& -\\
  
\midrule
2u 
& GNN-QE& 47.5 & - & - & -& -& -& -& -& -& -& 47.5& -\\
& ULTRA-query& 28.1 & - & - & -& -& -& -& -& -& -& 28.1& -\\
& ConE& 39.9 & - & - & -& -& -& -& -& -& -& 39.9& -\\
& CQD& 46.3 & - & - & -& -& -& -& -& -& -& 46.3& -\\
  & CQD-hybrid& 46.4 & - & - & -& -& -& -& -& -& -& 46.4& -\\
& QTO& 49.1 & - & - & -& -& -& -& -& -& -& 49.1& -\\
& CLMPT& 50.0 & - & - & -& -& -& -& -& -& -& 50.0& -\\
  
\midrule
2u1p 
& GNN-QE& 9.8 & 11.4 & - & -& -& -& -& -& -& -& 42.9& 5.2\\
& ULTRA-query& 9.5 & 12.3 & - & -& -& -& -& -& -& -& 36.9& 5.1\\
& ConE& 8.8 & 10.3 & - & -& -& -& -& -& -& -& 21.9& 6.3\\
& CQD& 18.5 & 18.8 & - & -& -& -& -& -& -& -& 53.5& 13.4\\
  & CQD-hybrid& 19.3 & 20.5 & - & -& -& -& -& -& -& -& 63.7& 13.1\\
& QTO& 18.9 & 21.1 & - & -& -& -& -& -& -& -& 58.2& 11.7\\
  & CLMPT& 22.0 & 17.6 & - & -& -& -& -& -& -& -& 29.7& 19.1\\
  
\bottomrule
\end{tabular}
}
\label{tab:strat-NELL-bal}

\end{table}
\begin{table}[ht]
\centering
\caption{\textbf{Stratified analysis on \icewsH}}
\scalebox{.63}{
\begin{tabular}{crcccccccccccc}
\toprule
Query type & method &overall& 1p & 2p & 3p & 4p & 2i & 3i & 4i & 1p2i & 2i1p &2u& 2u1p\\
\midrule
1p 
& GNN-QE& 12.2 & 12.2 & - & -& -& -& -& -& -& -& -& -\\
& ULTRA-query& 6.3 & 6.3 & - & -& -& -& -& -& -& -& -& -\\
& ConE& 3.5 & 3.5 & - & -& -& -& -& -& -& -& -& -\\
& CQD& 16.6 & 16.6 & - & -& -& -& -& -& -& -& -& -\\
  & CQD-hybrid& 16.6 & 16.6 & - & -& -& -& -& -& -& -& -& -\\
& QTO& 16.6 & 16.6 & - & -& -& -& -& -& -& -& -& -\\
& CLMPT& 4.7 & 4.7 & - & -& -& -& -& -& -& -& -& -\\
  
\midrule
2p 
& GNN-QE& 0.9 & 2.2 & 0.8 & -& -& -& -& -& -& -& -& -\\
& ULTRA-query& 1.2 & 2.1 & 1.1 & -& -& -& -& -& -& -& -& -\\
& ConE& 0.9 & 1.9 & 0.9 & -& -& -& -& -& -& -& -& -\\
& CQD& 2.6 & 5.4 & 2.4 & -& -& -& -& -& -& -& -& -\\
  & CQD-hybrid& 2.6 & 10.0 & 2.0 & -& -& -& -& -& -& -& -& -\\
& QTO& 2.6 & 8.4 & 2.2 & -& -& -& -& -& -& -& -& -\\
& CLMPT& 0.8 & 0.3 & 0.8 & -& -& -& -& -& -& -& -& -\\
  
\midrule
3p 
& GNN-QE& 0.5 & 1.0 & 1.2 & 0.4& -& -& -& -& -& -& -& -\\
& ULTRA-query& 1.2 & 2.1 & 1.5 & 1.2& -& -& -& -& -& -& -& -\\
& ConE& 0.9 & 1.5 & 1.4 & 0.9& -& -& -& -& -& -& -& -\\
& CQD& 1.5 & 2.6 & 3.2 & 1.4& -& -& -& -& -& -& -& -\\
  & CQD-hybrid& 1.5 & 9.8 & 4.4 & 1.2& -& -& -& -& -& -& -& -\\
& QTO& 1.4 & 6.2 & 3.7 & 1.2& -& -& -& -& -& -& -& -\\
& CLMPT& 0.1 & 0.2 & 0.1 & 0.1& -& -& -& -& -& -& -& -\\
\midrule
4p 
& GNN-QE& 0.4 & 1.6 & 1.4 & 1.0& 0.3& -& -& -& -& -& -& -\\
& ULTRA-query& 0.8 & 2.9 & 1.9 & 0.9& 0.7& -& -& -& -& -& -& -\\
& ConE& 0.6 & 1.8 & 1.8 & 0.9& 0.6& -& -& -& -& -& -& -\\
& CQD& 1.1 & 1.6 & 1.5 & 1.4& 1.1& -& -& -& -& -& -& -\\
  & CQD-hybrid& 0.9 & 31.4 & 39.3 & 20.1& 0.7& -& -& -& -& -& -& -\\
& QTO& 1.1 & 14.3 & 5.4 & 2.5& 0.7& -& -& -& -& -& -& -\\ 
& CLMPT& 0.1 & 0.1 & 0.1 & 0.1& 0.1& -& -& -& -& -& -& -\\ 
\midrule
2i 
& GNN-QE& 16.1 & 30.0 & - & -& 2.4& -& -& -& -& -& -& -\\
& ULTRA-query& 7.0 & 12.3 & - & -& 2.1& -& -& -& -& -& -& -\\
& ConE& 1.2 & 1.9 & - & -& 0.8& -& -& -& -& -& -& -\\
& CQD& 13.0 & 22.1 & - & -& 4.3& -& -& -& -& -& -& -\\
  & CQD-hybrid& 15.0 & 27.3 & - & -& 3.3& -& -& -& -& -& -& -\\
& QTO& 15.7 & 28.3 & - & -& 3.4& -& -& -& -& -& -& -\\
& CLMPT& 12.0 & 20.5 & - & -& 2.3& -& -& -& -& -& -& -\\
  
\midrule
3i 
& GNN-QE& 26.5 & 58.5 & - & -& -& 19.3& 2.0& -& -& -& -& -\\
& ULTRA-query& 11.7 & 25.9 & - & -& -& 8.0& 1.6& -& -& -& -& -\\
& ConE& 0.5 & 1.2 & - & -& -& 0.6& 0.2& -& -& -& -& -\\
& CQD& 19.5 & 42.3 & - & -& -& 13.4& 3.2& -& -& -& -& -\\
  & CQD-hybrid& 25.6 & 58.5 & - & -& -& 15.8& 2.7& -& -& -& -& -\\
& QTO& 24.7 & 56.7 & - & -& -& 16.3& 2.5& -& -& -& -& -\\
& CLMPT& 23.0 & 44.0 & - & -& -& 21.4& 2.9& -& -& -& -& -\\

\midrule
4i
& GNN-QE& 34.0 & 75.0 & - & -& -& 44.9& 13.9& 1.6& -& -& -& -\\
& ULTRA-query& 15.9 & 37.1 & - & -& -& 19.1& 6.0& 1.0& -& -& -& -\\
& ConE& 0.3 & 0.9 & - & -& -& 0.4& 0.2& 0.1& -& -& -& -\\
& CQD& 24.0 & 54.2 & - & -& -& 28.8& 10.2& 2.3& -& -& -& -\\
  & CQD-hybrid& 33.2 & 78.9 & - & -& -& 39.4& 11.8& 2.1& -& -& -& -\\
& QTO& 31.5 & 73.6 & - & -& -& 38.0& 11.9& 18.6& -& -& -& -\\
& CLMPT& 31.0 & 60.3 & - & -& -& 41.5& 18.4& 2.7& -& -& -& -\\
\midrule

1p2i 
& GNN-QE& 19.1 & 62.5 & 30.1 & -& -& 2.5& -& -& 2.3& -& -& -\\
& ULTRA-query& 8.8 & 19.6 & 13.2 & -& -& 28& -& -& 2.2& -& -& -\\
& ConE& 1.2 & 2.6 & 1.5 & -& -& 1.0& -& -& 0.8& -& -& -\\
& CQD& 17.1 & 38.5 & 25.8 & -& -& 50.1& -& -& 4.4& -& -& -\\
  & CQD-hybrid& 17.5 & 66.9 & 26.2 & -& -& 4.1& -& -& 3.5& -& -& -\\
& QTO& 18.4 & 59.3 & 28.0 & -& -& 4.4& -& -& 3.6& -& -& -\\
& CLMPT& 9.7 & 8.5 & 15.1 & -& -& 1.0& -& -& 2.3& -& -& -\\
  
\midrule
2i1p 
& GNN-QE& 3.5 & 4.5 & 3.0 & -& -& 5.7& -& -& -& 3.4& -& -\\
& ULTRA-query& 1.3 & 2.3 & 1.2 & -& -& 1.6& -& -& -& 1.2& -& -\\
& ConE& 1.6 & 2.4 & 1.6 & -& -& 2.4& -& -& -& 1.3& -& -\\
& CQD& 6.9 & 6.1 & 6.0 & -& -& 11.4& -& -& -& 6.8& -& -\\
  & CQD-hybrid& 5.8 & 9.2 & 4.6 & -& -& 14.8& -& -& -& 5.0& -& -\\
& QTO& 6.2 & 10.3 & 5.5 & -& -& 13.1& -& -& -& 5.1& -& -\\
& CLMPT& 2.1 & 0.5 & 1.6 & -& -& 0.4& -& -& -& 2.3& -& -\\
  
\midrule
2u 
& GNN-QE& 7.6 & - & - & -& -& -& -& -& -& -& 7.6& -\\
& ULTRA-query& 3.3 & - & - & -& -& -& -& -& -& -& 3.3& -\\
& ConE& 1.1 & - & - & -& -& -& -& -& -& -& 1.1& -\\
& CQD& 6.8 & - & - & -& -& -& -& -& -& -& 6.8& -\\
  & CQD-hybrid& 6.8 & - & - & -& -& -& -& -& -& -& 6.8& -\\
& QTO& 6.7 & - & - & -& -& -& -& -& -& -& 6.7& -\\
& CLMPT& 2.7 & - & - & -& -& -& -& -& -& -& 2.7& -\\
  
\midrule
2u1p 
& GNN-QE& 1.1 & 0.8 & - & -& -& -& -& -& -& -& 4.7& 0.8\\
& ULTRA-query& 1.2 & 1.5 & - & -& -& -& -& -& -& -& 2.0& 1.0\\
& ConE& 0.9 & 1.0 & - & -& -& -& -& -& -& -& 1.3& 0.8\\
& CQD& 5.9 & 3.2 & - & -& -& -& -& -& -& -& 11.6& 5.6\\
  & CQD-hybrid& 5.6 & 2.7 & - & -& -& -& -& -& -& -& 17.8& 4.5\\
& QTO& 4.9 & 3.3 & - & -& -& -& -& -& -& -& 17.8& 3.7\\
& CLMPT& 2.2 & 0.3 & - & -& -& -& -& -& -& -& 0.3& 2.5\\

\bottomrule
\end{tabular}
}
\label{tab:strat-ICEWS18-bal}

\end{table}

\begin{table}[ht]
\caption{\textbf{Also for queries involving negation, there is no clear SoTA method for the new benchmarks.} The MRR on the new benchmarks is significantly lower than the old ones. For example, for 3in queries on \fbnewH, QTO has an MRR of only 3.1, while for \fbnew it was 16.1. 
}

\small
    \centering
    \begin{tabular}{crccccccccccc}
    \toprule%
     \multicolumn{4}{r}{\fbnewH} & \multicolumn{4}{r}{\nellH}& \multicolumn{4}{r}{\icewsH}\\
     \cmidrule(lr){3-5}\cmidrule(lr){7-9}\cmidrule(lr){11-13}
     Query type & method & ovr 
& part-inf & full-inf & & ovr 
& par-inf & full-inf & & ovr 
& par-inf & full-inf\\

     \midrule
\multirow{4}{*}{2in} %
  & GNN-QE& 6.8&-&6.8&&5.5&-&5.5&&4.5&-&4.5\\
& ULTRAQ& 5.3&-&5.3&&4.5&-&4.5&&2.3&-&2.3\\
& CQD& 3.3&-&3.3&&4.2&-&4.2&&1.5&-&1.5\\
& CQD-hybrid& 4.7&-&4.7&&5.1&-&5.1&&1.7&-&1.7\\
& QTO& \textbf{10.6}&-&\textbf{10.6}&&\textbf{11.3}&-&\textbf{11.3}&&\textbf{4.9}&-&\textbf{4.9}\\
& ConE& 5.1&-&5.1&&4.6&-&4.6&&1.7&-&1.7\\
& CLMPT& 5.4&-&5.4&&5.2&-&5.2&&1.2&-&1.2\\
  \midrule
  \multirow{4}{*}{3in} %
  & GNN-QE& \textbf{6.5}&9.0&1.0&&\textbf{6.4}&7.2&1.0&&6.9&12.6&1.4\\
& ULTRAQ& 5.5&7.5&0.6&&5.9&6.2&0.9&&4.8&7.1&2.4\\
& CQD& 2.6&5.9&\textbf{2.5}&&1.5&6.0&1.3&&2.9&4.0&1.6\\
& CQD-hybrid& 1.6&8.4&1.2&&1.2&7.9&0.9&&4.0&6.5&1.3\\
& QTO& 3.1&13.9&2.3&&2.3&11.8&1.8&&\textbf{8.7}&14.5&\textbf{2.8}\\
& ConE& 4.9&6.4&2.0&&6.0&5.8&\textbf{2.8}&&2.9&3.3&2.0\\
& CLMPT& 2.3&9.3&1.7&&2.4&5.7&2.1&&2.1&3.4&0.9\\
  \midrule
  \multirow{4}{*}{2pi1pn} %
  & GNN-QE& \textbf{3.7}&4.0&1.4&&\textbf{5.8}&7.0&1.6&&0.9&1.0&0.5\\
& ULTRAQ& 2.6&2.8&1.3&&4.3&5.2&1.7&&\textbf{1.2}&1.3&0.9\\
& CQD& 0.6&1.3&0.5&&1.5&3.2&1.4&&0.2&0.7&0.1\\
& CQD-hybrid& 1.0&2.7&0.9&&1.4&5.9&1.1&&0.3&1.0&0.2\\
& QTO& 2.0&7.3&\textbf{1.7}&&3.1&9.4&\textbf{2.6}&&\textbf{1.2}&3.4&\textbf{1.0}\\
& ConE& 2.9&2.7&1.5&&3.7&4.9&2.0&&1.1&1.3&\textbf{1.0}\\
& CLMPT& 1.6&2.9&1.6&&2.2&4.3&2.1&&0.1&0.1&0.1\\
  \midrule
  \multirow{4}{*}{2nu1p} %
  & GNN-QE& \textbf{5.0}&-&\textbf{5.0}&&3.3&-&3.3&&\textbf{3.5}&-&\textbf{3.5}\\
& ULTRAQ& 3.7&-&3.7&&2.7&-&2.7&&2.2&-&2.2\\
& CQD& 4.2&-&4.2&&3.6&-&3.6&&2.7&-&2.7\\
& CQD-hybrid& 2.4&-&2.4&&2.9&-&2.9&&2.2&-&2.2\\
& QTO& 3.3&-&3.3&&\textbf{4.3}&-&\textbf{4.3}&&3.0&-&3.0\\
& ConE& 3.3&-&3.3&&2.7&-&2.7&&0.9&-&0.9\\
& CLMPT& 3.8&-&3.8&&3.1&-&3.1&&1.0&-&1.0\\
  \midrule
  \multirow{4}{*}{2in1p} %
  & GNN-QE& 3.3&2.4&1.4&&4.4&4.6&2.1&&0.8&1.0&0.5\\
& ULTRAQ& 2.2&2.2&1.2&&3.6&3.7&1.6&&\textbf{1.6}&2.2&\textbf{1.4}\\
& CQD& 1.2&1.3&1.2&&2.6&3.5&2.6&&0.9&0.2&0.9\\
& CQD-hybrid& 1.3&1.1&1.2&&2.4&4.1&2.3&&1.1&1.9&1.0\\
& QTO& 1.5&5.4&1.3&&2.4&7.9&1.9&&0.9&2.5&0.9\\
& ConE& \textbf{3.6}&2.8&2.2&&\textbf{6.4}&7.0&\textbf{3.9}&&1.3&2.3&1.2\\
& CLMPT& 2.5&3.4&\textbf{2.4}&&4.0&6.3&3.8&&0.2&0.1&0.2\\
  \bottomrule
   
\end{tabular}
    \label{tab:neg_prop_MRR}
\end{table}
\subsection{Statistical tests}
{
As an illustrative case, we analyze 2p queries from \fbnewH to assess whether the observed differences in model performance are statistically significant, despite similar MRR in \cref{tab:new-bench+h-MRR}. We conduct an all-vs-all paired Mann–Whitney U test on the reciprocal ranks across all models for each 2p QA pair. The results in \cref{tab:stat-tests-2p} reveal strong statistical differences in most pairwise comparisons, indicating that models rank correct answers differently even when aggregate metrics are close. However, a few model comparisons—such as ConE vs. CLMPT and CQD vs. GNNQE—exhibit statistical similarity, suggesting comparable ranking behavior on 2p queries. To examine whether the similarity is limited to specific query types, we apply the same test across all QA pairs of \fbnewH from all query types. The resulting p-values are effectively zero for all comparisons, confirming that significant performance differences are widespread, though limited cases of similarity may occur between some models on some specific query types.

}
\clearpage
\newpage
\section{Queries with four hops}\label{4p4iform}
Logical formulation of the 4p and 4i queries we introduced in the benchmarks \fbnewH, \nellH, \icewsH
\begin{equation} \label{4p}
\tag{4p}
\begin{aligned}
    ?T: &\ \exists \varQuantified_1, \varQuantified_2, \varQuantified_3 . 
    (\anchorA_1, \relR_1, \varQuantified_1) \land (\varQuantified_1, \relR_2, \varQuantified_2) \land \\
        &\ (\varQuantified_2, \relR_3, \varQuantified_3) \land (\varQuantified_3, \relR_4, \varTarget),
\end{aligned}
\end{equation}

\begin{equation} \label{4i}
\tag{4i}
\begin{aligned}
    ?T: &\ (\anchorA_1, \relR_1, \varTarget) \land
    (\anchorA_2, \relR_2, \varTarget) \land \\
        &\ (\anchorA_3, \relR_3, \varTarget) \land
    (\anchorA_4, \relR_4, \varTarget).
\end{aligned}
\end{equation}
\section{Hyperparameters}\label{app:hyperparams}
In this section, we detail the hyperparameters used for each dataset and model. Old and new benchmarks, the generation scripts, and the implementation of CQD-hybrid  are included in our official repo.\footnote{https://github.com/april-tools/is-cqa-complex}
\subsection{Old benchmarks}\label{app:hyperparams_old}
\paragraph{GNN-QE} We did not tune hyperparameters, but re-used the ones provided in the official repo.\footnote{https://github.com/DeepGraphLearning/GNN-QE/tree/master/config}
\paragraph{ULTRA}
We did not tune hyperparameters, but re-used the ones provided in the official repo.\footnote{https://github.com/DeepGraphLearning/ULTRA/tree/main/config/ultraquery}
\paragraph{CQD}
As mentioned in \cref{sec:performance}, we re-used the pre-trained link predictor \citep {DBLP:journals/jmlr/TrouillonDGWRB17} provided by the authors.\footnote{https://github.com/Blidge/KGReasoning/} However, we tuned CQD-specific hyperparameters, namely the CQD beam ``k'', ranging from [2,512] and the t-norm type being ``prod'' or ``min''
In \cref{tab:hyper_CQD_old} we provide the hyperparameter selection for the old benchmarks \fbnew and \nell. Also note that we normalize scores with min-max normalization. For negation, we always use standard negation. 
\paragraph{QTO}
We re-used the hyperparameters provided in the official repo. \footnote{https://github.com/bys0318/QTO/tree/main} For a fair comparison with CQD, CQD-hybrid, we used the same pre-trained link predictor.
\paragraph{CLMPT}
We did not tune hyperparameters but re-used the ones provided in the official repo.\footnote{https://github.com/qianlima-lab/clmpt} 

\begin{table}
    \centering
    \scalebox{0.7}{
    \begin{tabular}{lccccccccccccccccccccccccccc}
        \toprule
        Dataset & \multicolumn{2}{c}{2p} & \multicolumn{2}{c}{3p} & \multicolumn{2}{c}{2i} & \multicolumn{2}{c}{3i}& \multicolumn{2}{c}{1p2i}& \multicolumn{2}{c}{2i1p}& \multicolumn{2}{c}{2u}& \multicolumn{2}{c}{2u1p}& \multicolumn{2}{c}{2in}& \multicolumn{2}{c}{3in}& \multicolumn{2}{c}{2pi1pn}& \multicolumn{2}{c}{2nu1p}& \multicolumn{2}{c}{2in1p}\\
        \cmidrule(lr){1-1} \cmidrule(lr){2-3} \cmidrule(lr){4-5} 
        \cmidrule(lr){6-7} \cmidrule(lr){8-9} \cmidrule(lr){10-11} \cmidrule(lr){12-13} \cmidrule(lr){14-15} \cmidrule(lr){16-17} 
        \cmidrule(lr){18-19} \cmidrule(lr){20-21} 
        \cmidrule(lr){22-23} \cmidrule(lr){24-25} 
        \cmidrule(lr){26-27}
         & k & tn & k & tn & k & tn & k & tn & k & tn & k & tn & k & tn & k & tn & k & tn & k & tn & k & tn & k & tn & k & tn\\
        \midrule
        \fbnew & 512 & p & 8 & p & 128 & p & 128 & p & 256 & p & 64 & p & 512 & m & 512 & m & 2 & p& 2 & p& 512 & prod& 512 & prod& 512 & prod\\
        \nell & 512 & p & 2 & p & 2 & p & 2 & p & 256 & p & 256 & p & 512 & m & 512 & m & 2 & p& 2 & p& 512 & prod& 512 & prod& 512 & prod\\
        \bottomrule
    \end{tabular}}
    \caption{CQD hyperparameters old benchmarks. tn = tnorm. p=prod, m=min. Note that for 1p neither the k nor the tnorm are needed.}
    \label{tab:hyper_CQD_old}
\end{table}

\paragraph{ConE}
We did not tune hyperparameters but re-used the ones provided in the official repo.\footnote{https://github.com/MIRALab-USTC/QE-ConE/blob/main/scripts.sh} 
\paragraph{CQD-Hybrid}
For CQD-Hybrid, to make the comparison with CQD fair, we re-used the hyperparameters found for CQD and fixed an upper bound value for the CQD beam ``k'' to 512, even when there are more existing entities matching the existentially quantified variables, to match the upper bound of the ``k'' used for CQD. Additionally, the scores of the pre-trained link predictor are normalized between $[0,0.9]$ using min-max normalization, and a score of $1$ is assigned to the existing triples.
\subsection{New benchmarks}\label{app:hyperparams_new}
For \fbnewH and \nellH, we re-used the same models trained for the old benchmarks, and using the same hyperparameters presented in \cref{app:hyperparams_old}. The only exception is the hyperparameters for CQD and CQD-hybrid of 4p and 4i queries that were not tuned for the old benchmarks. We used k=2 and tnorm=prod for both query types in all benchmarks. Moreover, for \ref{2u} queries on \fbnewH we used tnorm=prod. 

Instead, for the new benchmark \icewsH we trained every model. The used hyperparameters for each model are presented in the following:
\paragraph{GNN-QE}
For GNN-QE, we tuned the following hyperparameters: (1) \textit{batchsize}, with values 8 or 48, and \textit{concat hidden} being True or False, while the rest are the same used for the old benchmarks and do not change across benchmarks.
For \icewsH the best hyperparameters are ``batchsize=48'' and ``concat hidden=True''.
\paragraph{ULTRAQ}
Being a zero-shot neural link predictor, we re-used the same checkpoint provided in the official repo, as for \ref{app:hyperparams_old}.
\paragraph{CQD}
We train ComplEx \cite{DBLP:journals/jmlr/TrouillonDGWRB17} link predictor with hyperparameters ``regweight'' 0.1 or 0.01, and batch size 1000 or 2000, with the best being, respectively 0.1 and 1000. Moreover, in \ref{tab:hyper_CQD_ICE18} are shown the hyperparameters for CQD on the new benchmark \icewsH.
\paragraph{QTO}
We re-used the link predictor used in CQD and CQD-Hybrid, and for QTO-specific hyperparameters, we used the same of \fbnew.
\begin{table}
    \centering
    \small
    \scalebox{0.7}{
    \begin{tabular}{lccccccccccccccccccccccccccccccc}
        \toprule
        Dataset & \multicolumn{2}{c}{2p} & \multicolumn{2}{c}{3p} & \multicolumn{2}{c}{2i} & \multicolumn{2}{c}{3i}& \multicolumn{2}{c}{1p2i}& \multicolumn{2}{c}{2i1p}& \multicolumn{2}{c}{2u}& \multicolumn{2}{c}{2u1p}& \multicolumn{2}{c}{4i}& \multicolumn{2}{c}{4p}& \multicolumn{2}{c}{2in}& \multicolumn{2}{c}{3in}& \multicolumn{2}{c}{2pi1pn}& \multicolumn{2}{c}{2nu1p}& \multicolumn{2}{c}{2in1p}\\
        \cmidrule(lr){1-1} \cmidrule(lr){2-3} \cmidrule(lr){4-5} \cmidrule(lr){6-7} \cmidrule(lr){8-9} \cmidrule(lr){10-11} \cmidrule(lr){12-13} \cmidrule(lr){14-15} \cmidrule(lr){16-17} \cmidrule(lr){18-19} \cmidrule(lr){20-21}
        \cmidrule(lr){22-23} \cmidrule(lr){24-25}
        \cmidrule(lr){26-27} \cmidrule(lr){28-29}
        \cmidrule(lr){30-31} 
         & k & tn & k & tn & k & tn & k & tn & k & tn & k & tn & k & tn & k & tn & k & tn & k & tn & k & tn & k & tn & k & tn & k & tn & k & tn\\
        \midrule
        \icewsH & 32 & p & 2 & p & 2 & p & 2 & p & 512 & p & 256 & p & 2 & m & 2 & m & 2 & p & 2 & p & 2 & p & 2 & p & 512 & prod & 512 & prod & 512 & prod \\
        \bottomrule
    \end{tabular}}
    \caption{CQD hyperparameters \icewsH benchmark. tn = tnorm. p=prod, m=min}
    \label{tab:hyper_CQD_ICE18}
\end{table}

\paragraph{CQD-Hybrid}
For CQD-Hybrid, as mentioned above, we re-used the hyperparameters used for CQD. Note that for 4p queries,  we fixed the ``k'' upper bound to 64 for memory constraints. 
\paragraph{ConE} 
We re-used %
the same hyperparameters of \nell for \icewsH.  

\paragraph{CLMPT} For CLMPT, we tuned the following hyperparameters: (1) learning rate, with values in [1e-5,5e-2,5e-3,5e-4,5e-5,,5e-6], (2) temp, with values in [0.1, 0.2]
We found the best hyperparameters to match the same found for \fbnew. Hence, we re-used them for \icewsH.

\section{Statistics}
\subsection{KG splits statistics}\label{app:kg_statistics}
The statistics, i.e., number of entities, relation names, training/validation/test links, of the knowledge graphs used in this paper are shown in \cref{tab:datasets statistics}.
\begin{table}[ht]
\centering
\caption{Statistics of knowledge graphs used to generate complex queries}
\label{tab:datasets statistics}
\scalebox{.83}{
\begin{tabular}{cccccccc}
\toprule
Dataset & Entities & Relation Names & Training Links & Validation Links & Test Links & Total Links \\
\midrule
\fbnew & 14,505 & 237 & 272,115 & 17,526 & 20,438 & 310,079 \\

\nell & 63,361 & 200 & 114,213 & 14,324 & 14,267 & 142,804 \\
\midrule
\icewsCQ (\textbf{\textit{new}}) & 20,840 & 250 & 213,304 & 25,048 & 24,689 & 263,041 \\
\bottomrule
\end{tabular}}

\end{table}

The KG splits of \nell and \fbnew are the same of the one used in \citep{DBLP:conf/nips/RenL20, DBLP:conf/iclr/RenHL20}. 
\subsection{New benchmark statistics}\label{app:new_bench_stats}
\subsubsection{\nellH and \fbnewH}\label{sec:stats_nf}
\paragraph{Training queries} were not changed from the one proposed by \citet{DBLP:conf/nips/RenL20, DBLP:conf/iclr/RenHL20}.
\paragraph{Validation and Test} queries are generated to ensure a balanced distribution of hardness across the sub-query types to which a given query can be reduced. As described in \cref{sec:new-benchmarks}, we generate queries such that each sub-query type contains exactly 10,000 
QA pairs, along with an additional 10,000 pairs corresponding to full-inference queries (i.e., non-reducible cases). To achieve this, we adapt the algorithm from \citep{DBLP:conf/iclr/RenHL20}, sampling queries until the required number of 
QA pairs for each sub-query type is reached.
For instance, for 2p queries, we continue sampling until we obtain 10,000 QA pairs reducible to 1p and 10,000 full-inference queries. When a query is sampled, all its corresponding answers are included in the benchmark until the target of 10,000 
QA pairs per subtype is met. If incorporating all answers would exceed this limit, a random subset is selected to ensure exactly 10,000 pairs. Similarly, once a sub-query type reaches the 10,000-pair threshold, any additional answers belonging to that category are discarded.

Only exceptions are \ref{1p}, \ref{2in}, and \ref{2nu1p}, which being already of full-inference type we kept the same as in \citep{DBLP:conf/nips/RenL20, DBLP:conf/iclr/RenHL20}. 

\subsubsection{ICEWS18+H}
\paragraph{Training queries} were generated using the same strategy used in \citep{DBLP:conf/nips/RenL20, DBLP:conf/iclr/RenHL20} , namely by (1) creating \ref{1p} queries by considering every triple in the training set, (2) retrieve the number $n$ of generated \ref{1p} queries, (3) generate $n$ queries for all training query types.
\paragraph{Validation and test queries} were generated using the same strategy detailed in \cref{sec:stats_nf}. Moreover, since we generated this benchmark from scratch, we also had to generate  \ref{1p} (which following \citep{DBLP:conf/nips/RenL20, DBLP:conf/iclr/RenHL20}, were generated for each triple in the validation and test split), and \ref{2in}, \ref{2nu1p} queries.